\documentclass[10pt]{IEEEtran} 
\usepackage{times}
\usepackage{hyperref}
\usepackage{url}

\usepackage{url}
\usepackage{amsmath}
\usepackage{amssymb}
\usepackage{graphicx}
\usepackage[]{caption,subfig}
\usepackage{mdwlist}
\usepackage[lined, boxed]{algorithm2e}
\usepackage{multirow}
\usepackage{amsthm}
\usepackage{color}
\usepackage{comment}

\graphicspath{{figs/}}

\title{Data Representation using the Weyl Transform}

\author{Qiang Qiu$^*$\thanks{$^*$Q. Qiu, R. Calderbank and G. Shapiro are with the Department of Electrical Engineering, Duke University, USA; email: {\tt\small  \{qiang.qiu, robert.calderbank, guillermo.sapiro\}@duke.edu; }}, Andrew Thompson$^{\dag}$\thanks{$^{\dag}$A. Thompson is with the Mathematical Institute, University of Oxford, UK; email: {\tt\small thompson@maths.ox.ac.uk}}, Robert Calderbank$^*$, and Guillermo Sapiro$^*$\thanks{Q. Qiu and A. Thompson have contributed equally to this work.}}

\newcommand{\RR}{\mathbb{R}}
\newcommand{\ZZ}{\mathbb{Z}}
\newcommand{\YY}{\mathbb{Y}}

\newtheorem{theorem}{Theorem}[section]
\newtheorem{lemma}[theorem]{Lemma}
\newtheorem{proposition}[theorem]{Proposition}

\begin{document}

\maketitle

\begin{abstract}
The Weyl transform is introduced as a rich framework for data representation. Transform coefficients are connected to the Walsh-Hadamard transform of multiscale autocorrelations, and different forms of dyadic periodicity in a signal are shown to appear as different features in its Weyl coefficients. The Weyl transform has a high degree of symmetry with respect to a large group of multiscale transformations, which allows compact yet discriminative representations to be obtained by pooling coefficients. The effectiveness of the Weyl transform is demonstrated through the example of textured image classification.
\end{abstract}


\section{Introduction}
\label{sec:intr}

Many signal processing tasks, such as detection, clustering, and classification, rely on representations which are invariant to a given group of transformations. Additional invariance to transformations which permute feature coefficients can often be achieved by pooling coefficients, for example Gabor wavelets~\cite{Gabor} and scattering transforms~\cite{scattering}.

Our focus here is not on new bases for representing signals, rather it is on the type of measurement that is fundamental to widely used algorithms for signal processing tasks. Thus, our focus is autocorrelation, and we show that autocorrelations can be calculated from trace inner products of covariance matrices with signed permutation matrices from the discrete Heisenberg-Weyl group. When a signal is transformed by an element of the (much larger) discrete symplectic group, these autocorrelations are fixed up to permutation and sign change. The symplectic group can be viewed as a discrete approximation to the full unitary group,
and because it is so large (the order is $2^{m^2}\cdot(2^{2m}-1)(2^{2m-2}-1)\ldots(2^2-1)$ for a signal of length $2^m$), it allows for great versatility in the design of pooling strategies. Our approach makes use of the power of autocorrelation for describing multiscale periodicity in signals, and we demonstrate through the example of texture classification that this framework is useful for signal representation.

The mapping between a signal and its autocorrelation coefficients based on the Heisenberg-Weyl group is called the \emph{Weyl transform}. This instance of the Weyl transform is a special case of a general framework for representation of operators in harmonic analysis. In radar the larger (continuous) framework is fundamental to the study of the radar ambiguity function~\cite{finite_HW,finite_HW_radar,harmonic}. The binary Heisenberg-Weyl group studied here plays an important role in coding theory~\cite{Z4}.
In this paper we make a new connection to signal analysis by describing the Weyl transform in terms of the Walsh-Hadamard transform~\cite{hadamard_transform} of binary autocorrelations.

Autocorrelation is known to be a powerful tool for detecting periodicity while imposing shift invariance, and has been used in a variety of signal processing tasks, including but not restricted to speech coding~\cite{speech}, pitch estimation~\cite{pitch} and noise removal~\cite{noise_removal}.
Applications in image processing include character recognition, face detection, texture classification, and pattern recognition~\cite{texture_class,face_detection,character,pattern_recog}. Autocorrelation can be cyclic, or dyadic~\cite{binary_face}, the latter being especially well suited to representing multiscale texture. The Walsh-Hadamard Transform (WHT) is also well suited to periodic signals since it captures binary harmonics of data, being the binary counterpart of the Discrete Fourier Transform (DFT). The WHT can be combined with the Wiener-Khintchine convolution theorem to provide a fast method for calculating dyadic autocorrelation~\cite{binary_calculation}, and the windowed WHT has been used for pattern matching of images~\cite{hadamard_patterns}. While all of the aforementioned previous work involving autocorrelation and the WHT is to some extent heuristic, we develop for the first time a rigorous mathematical theory of invariance which combines the two notions.
\begin{figure} [t]
\centering
 \subfloat[\emph{Jeans}] {\includegraphics[angle=0, height=0.1\textwidth, width=.1\textwidth]{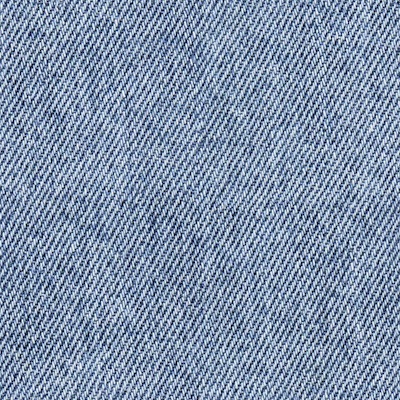}}
 \subfloat[\emph{5 Jeans patches}] {\includegraphics[angle=0, height=0.08\textwidth, width=.38\textwidth]{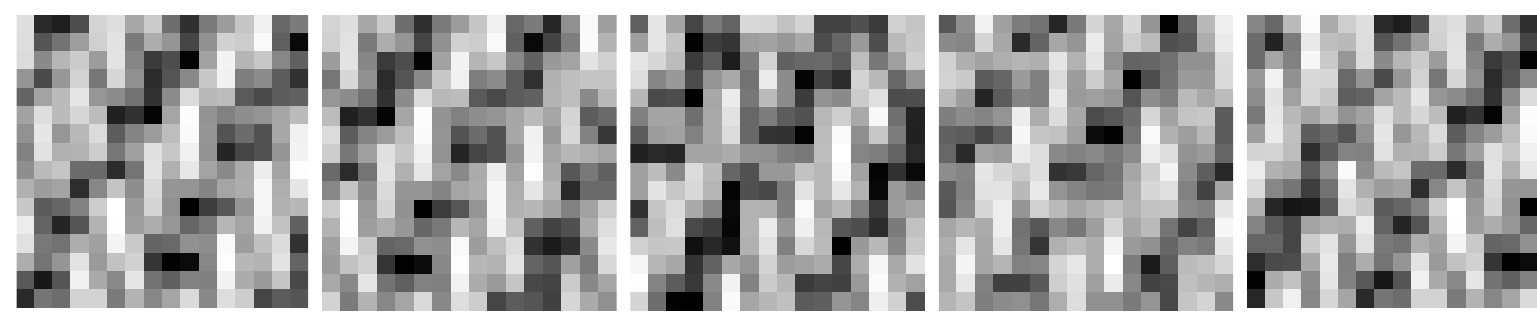}}\\
      \subfloat[\emph{Cotton}] {\includegraphics[angle=0, height=0.1\textwidth, width=.1\textwidth]{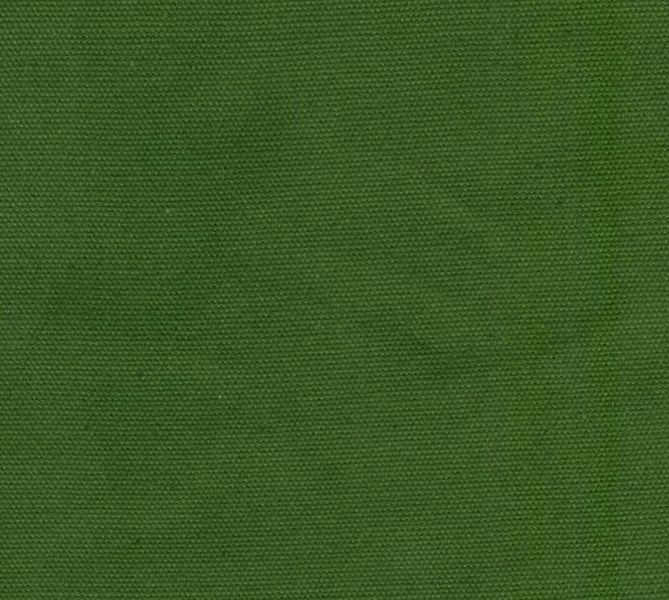}}
    \subfloat[\emph{5 Cotton patches}] {\includegraphics[angle=0, height=0.08\textwidth, width=.38\textwidth]{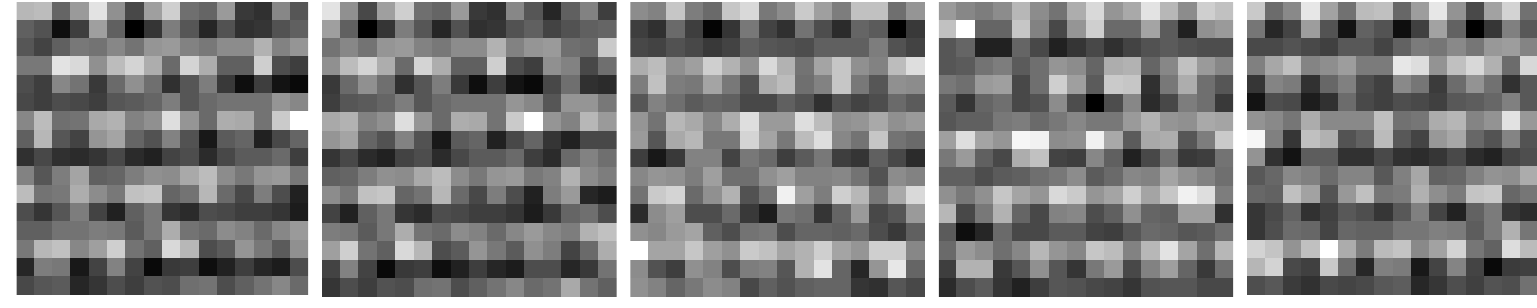}} \\
  \subfloat[\emph{Weyl transforms of (b)}] {\includegraphics[angle=0, height=0.08\textwidth, width=.46\textwidth]{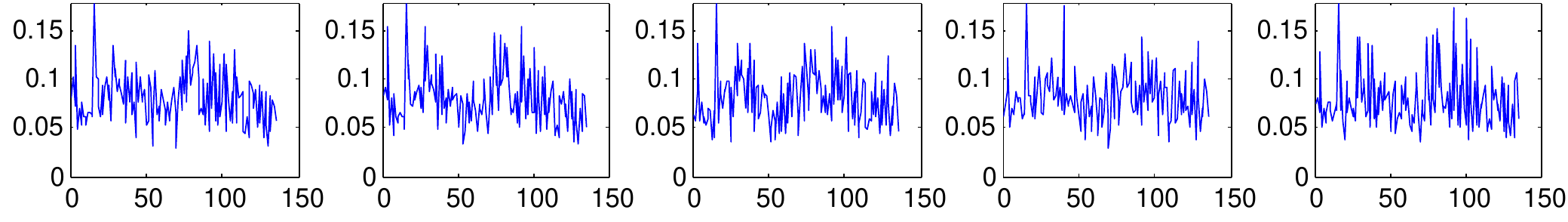}} \\
  \subfloat[\emph{Weyl transforms of (d)}] {\includegraphics[angle=0, height=0.08\textwidth, width=.46\textwidth]{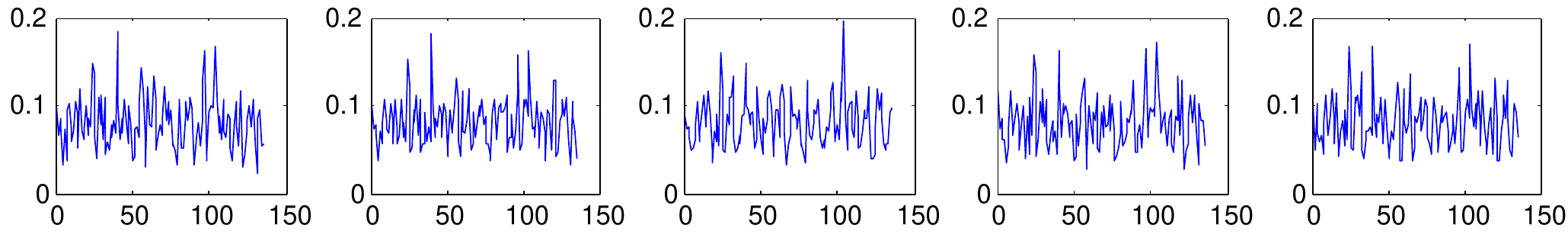}}\\
    \subfloat[\emph{Similarity}] {\includegraphics[angle=0, height=0.14\textwidth, width=.14\textwidth]{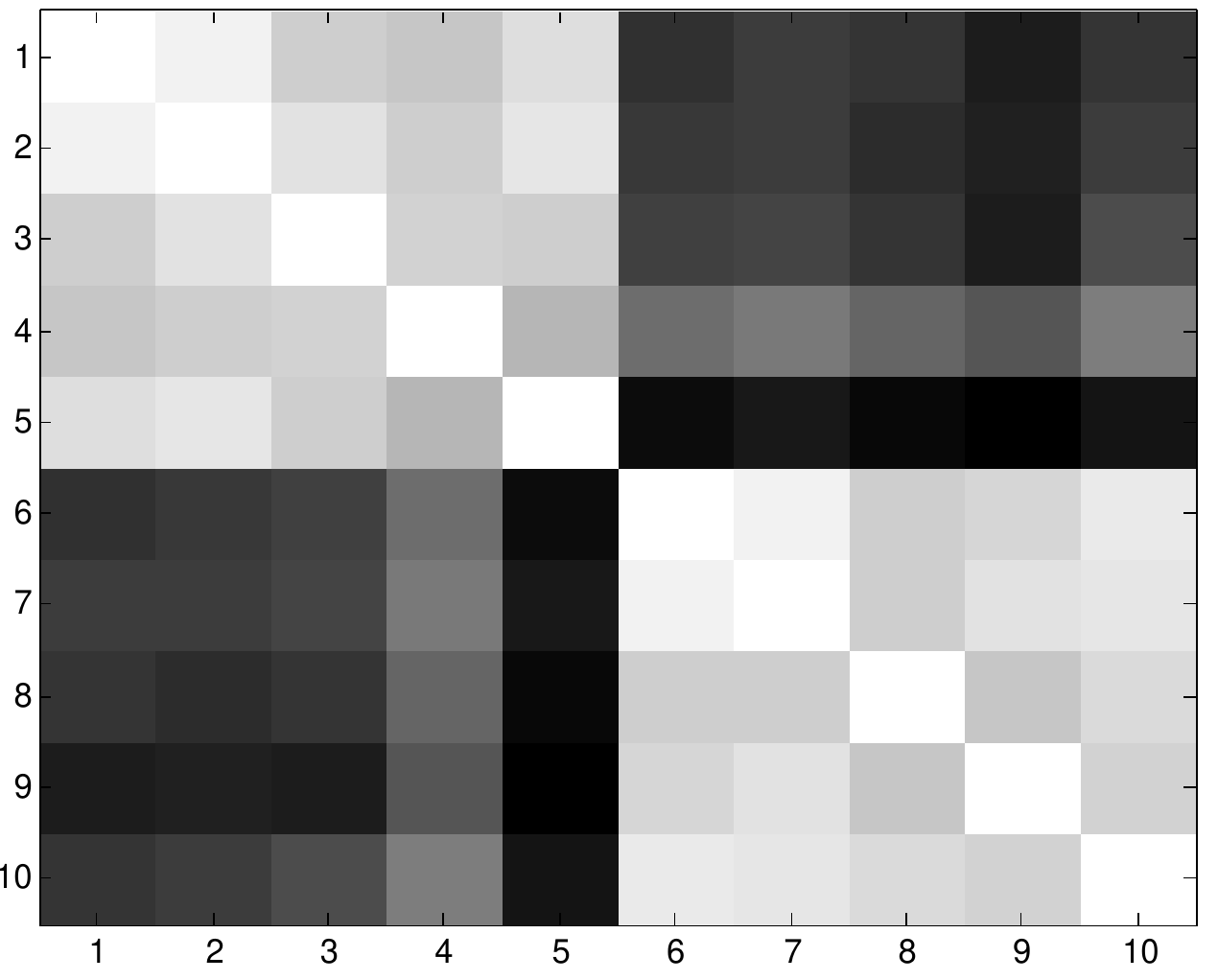}}
\caption{The Weyl transform examples. (b) and (d) are
 five $16\times 16$ patches randomly sampled from two textures (a) and (c), respectively.
(e) and (f) are the Weyl transforms of (b) and (d); see Section~\ref{sec:thm} for details.
(g) shows the similarity matrix for the Weyl transforms of patches from \emph{Jeans} followed by \emph{Cotton}, where darker colors indicate lower similarity.
As shown in (g),
patches sampled from the same texture share similar Weyl transforms, even though they
exhibit obvious dissimilarity, e.g., due to translation.
}
\label{fig:intro}
\end{figure}

Fig.~\ref{fig:intro} illustrates how effectively the Weyl transform is able to encode differences in texture. It displays five randomly sampled 16x16 patches from each of two different textures. Each patch is broken down into smaller $4\times 4$ sub-patches, their Weyl transforms calculated, and their absolute values aggregated over all sub-patches. Fig.~\ref{fig:intro}(g) displays a similarity matrix where darker entries indicate greater Euclidean distance between Weyl transforms of different patches. Adjacent patches form clusters that clearly separate the two textures.

Section~\ref{sec:thm} introduces the Weyl transform and its invariance property, namely that the absolute values of Weyl coefficients are invariant to certain symmetries of the signal (Theorem~\ref{thm:invariance}). We also establish a connection between the Weyl transform and the WHT of dyadic autocorrelations (Theorem~\ref{thm:hadamard}).
In Section~\ref{sec:exa}, we illustrate the theory with examples of real-world textures. We demonstrate the versatility of the Weyl transform by pooling coefficients to obtain features with additional invariance to $90^{\circ}$ rotation and cyclic translations. We also describe a supervised learning example in which training data is used to select the Weyl coefficients that are most significant in distinguishing the classes. For both examples, we compare our approach with three state-of-the-art image representations: Gabor wavelets~\cite{Gabor}, HOG~\cite{HOG} and LBP~\cite{LBP}. On the examples tested, the Weyl transform regularly outperforms these other methods while using a significantly shorter feature vector. Full proofs of all results can be found in Section~\ref{proofs}.

\section{Weyl transform theory}
\label{sec:thm}

We now introduce some new fundamental theory about the Weyl transform. We first give a summary of the theory in Section~\ref{sec:summary}, in which we define the Weyl transform and give a crucial result about its invariance to certain multiscale transformations. The Weyl transform consists of inner products with matrices from the binary Heisenberg-Weyl group, and we describe this group of matrices more fully in Section~\ref{sec:HWG}. In Section~\ref{sec:hadamard}, we make a connection between the Weyl transform and the Walsh-Hadamard Transform. Note that \cite{Z4,finite_HW,finite_HW_radar} show material related to the discrete Weyl transform, but presented in a different context, less suited to the data and applications described here. 

\subsection{The discrete Weyl transform}\label{sec:summary}

\begin{figure} [h!]
\centering
{\label{fig:pose29} \includegraphics[angle=0, height=0.033\textwidth, width=.5\textwidth]{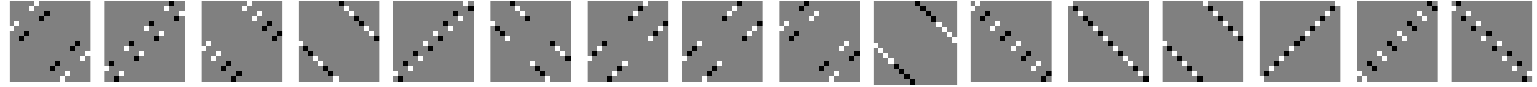}}
\caption{Examples of the matrices $D(a,b)$ for $m=4$: -1 (black), 0 (grey), 1 (white).}
\label{fig:signed_perm}
\end{figure}

Given a vectorized signal $y\in\RR^{2^m}$, we define its \emph{Weyl coefficients} $\omega_{a,b}(y)$ to be
$$\omega_{a,b}(y):=\frac{1}{2^{m/2}}\mathrm{Tr}[yy^T\cdot D(a,b)],$$
where $a=(a_{m-1}\ldots a_0)^T$ and $b=(b_{m-1}\ldots b_0)^T$ are binary $m$-tuples, and where the $\{D(a,b)\}$ are multiscale signed permutation matrices from the binary Heisenberg-Weyl group (see Section~\ref{sec:HWG} for more details). We will denote the set of binary $m$-tuples by $\ZZ_2^m$. Figure~\ref{fig:signed_perm} gives examples of matrices $D(a,b)$ for $m=4$.

The mapping from covariance matrices $yy^T\in\RR^{2^m\times 2^m}$ to vectors of Weyl coefficients in $\RR^{2^{2m}}$ is an isometry, which we will refer to as the \emph{Weyl transform}. A main result of this paper is that, whenever a $D(a,b)$ matrix is applied to a signal vector $y$, the magnitudes of its Weyl coefficients are unchanged. The Weyl transform is therefore invariant to a large class of multiscale signed permutations, a desirable property for a representation designed to detect periodicity.\vspace{3pt}

\begin{theorem}[\textbf{Weyl transform invariance property}]\label{thm:invariance}
Let $\{\omega_{a,b}(y)\}$ be the Weyl coefficients of $y\in\RR^{2^m}$. If $y'=D(a',b')y$ for some $(a',b')\in\ZZ_2^m$, then, for all $(a,b)\in\ZZ_2^m$ such that $a^T b=0$\footnote{The result in fact holds for all $(a,b)$, but we make the assumption $a^T b=0$ to simplify some of the analysis.},
$|\omega_{a,b}(y')|=|\omega_{a,b}(y)|$.
\end{theorem}
\textbf{Proof:} See Section~\ref{invariance}.

\begin{figure} [t]
\centering
\subfloat[] {\includegraphics[angle=0, height=0.15\textwidth, width=.15\textwidth]{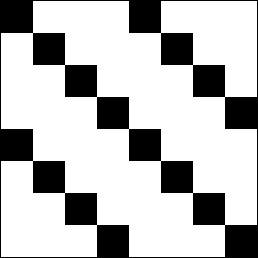}} \\
\subfloat[] {\includegraphics[angle=0, height=0.07\textwidth, width=.07\textwidth]{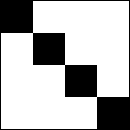} \hspace{0pt}
 \includegraphics[angle=0, height=0.07\textwidth, width=.07\textwidth]{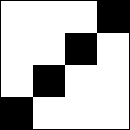} \hspace{0pt}
 \includegraphics[angle=0, height=0.07\textwidth, width=.07\textwidth]{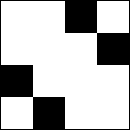} \hspace{0pt}
\includegraphics[angle=0, height=0.07\textwidth, width=.07\textwidth]{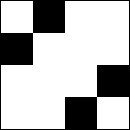}}\\
\subfloat[] {\includegraphics[angle=0, height=0.1\textwidth, width=.1\textwidth]{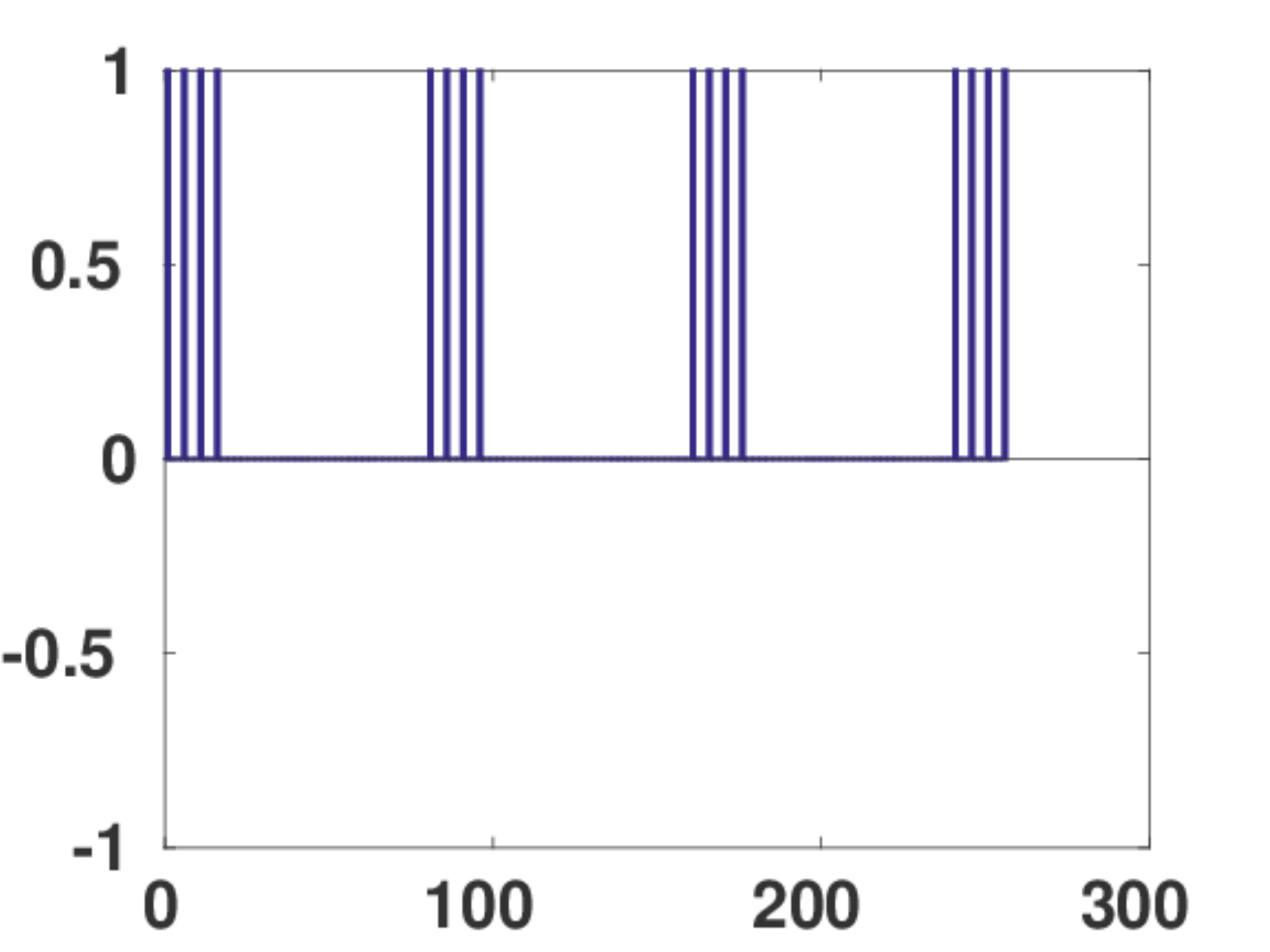} \hspace{0pt}
 \includegraphics[angle=0, height=0.1\textwidth, width=.1\textwidth]{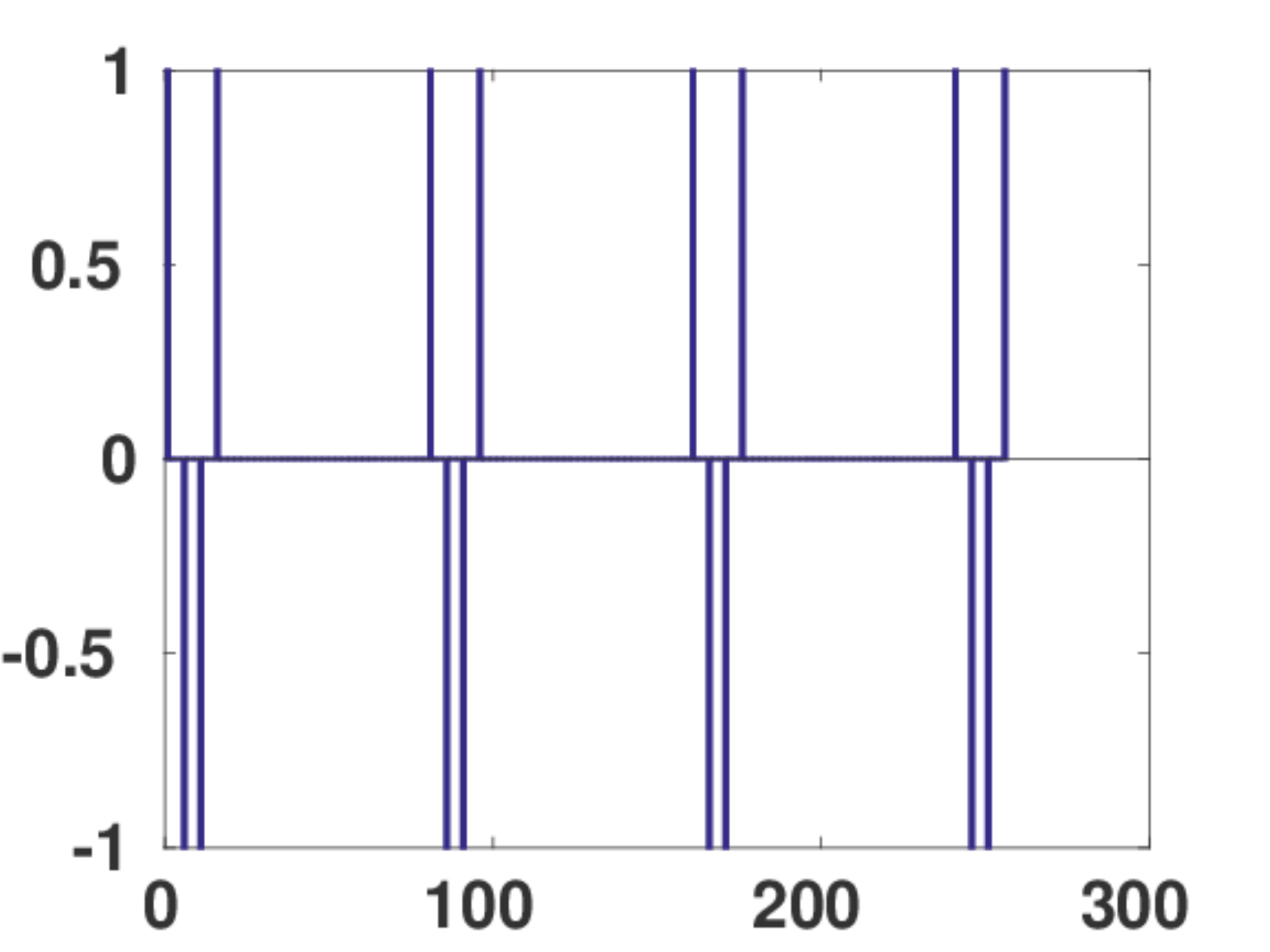} \hspace{0pt}
 \includegraphics[angle=0, height=0.1\textwidth, width=.1\textwidth]{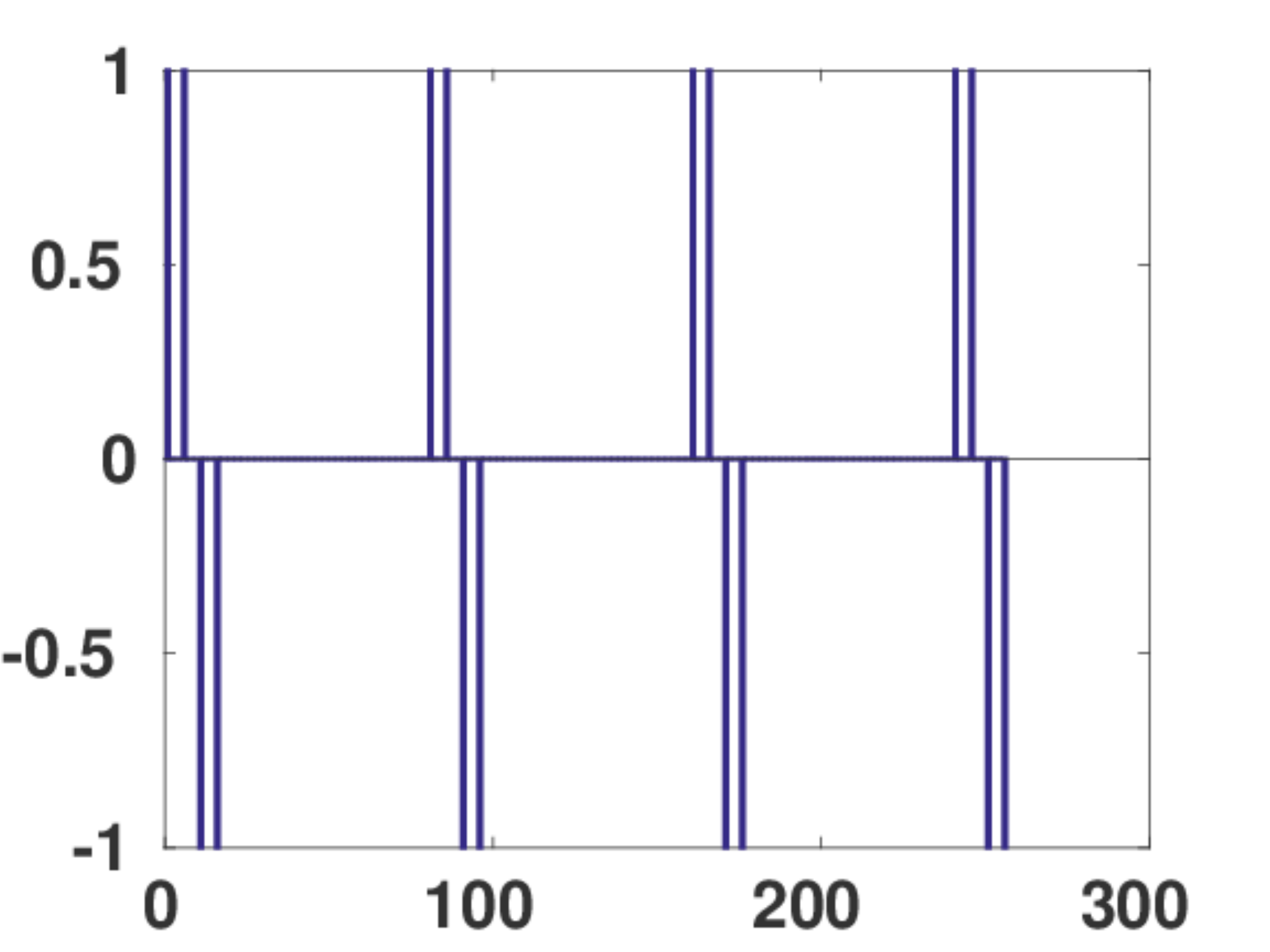} \hspace{0pt}
 \includegraphics[angle=0, height=0.1\textwidth, width=.1\textwidth]{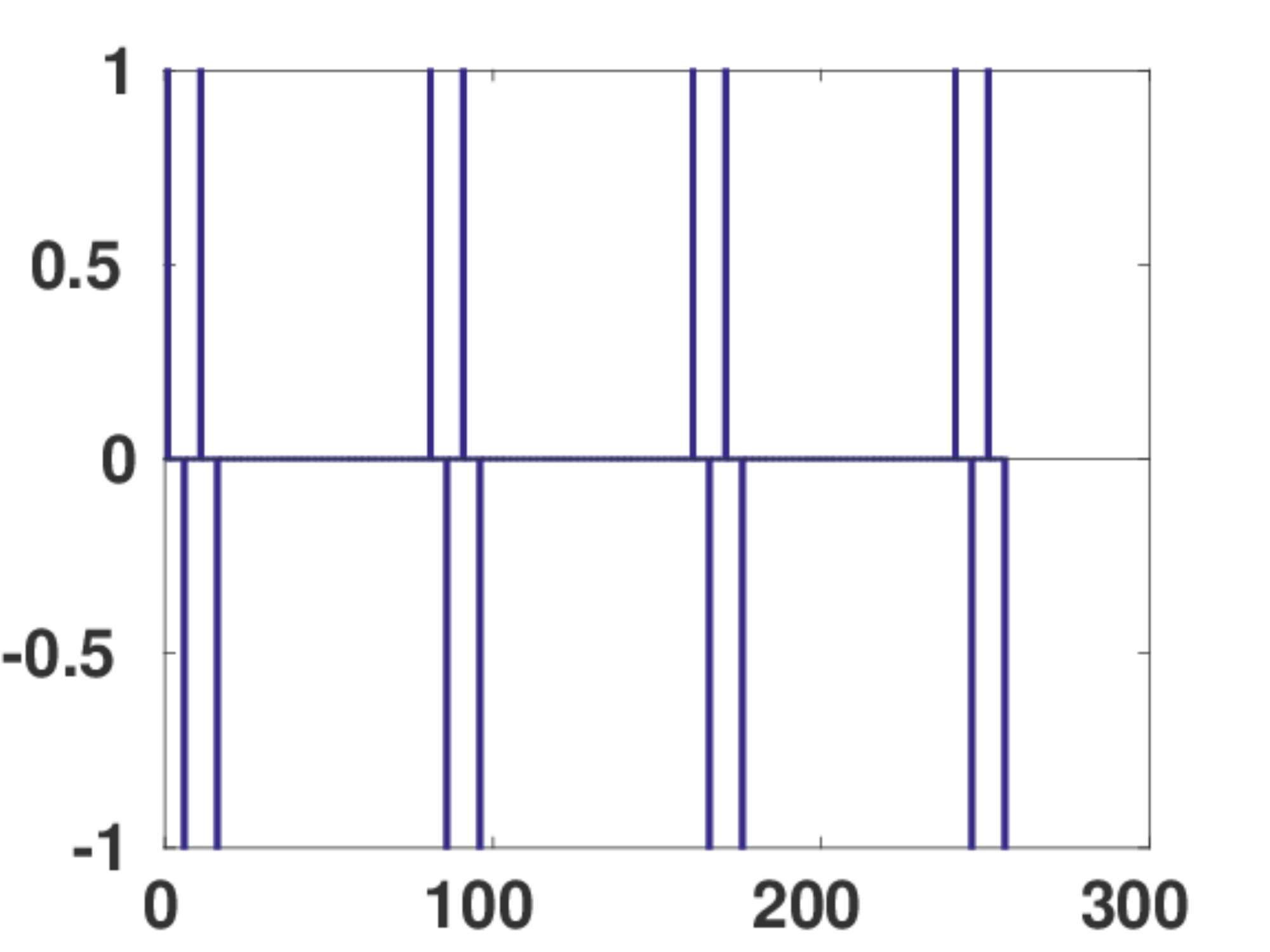}}
\caption{(a) A simple texture pattern; (b) Four subpatches taken at different positions and orientations; (c) The Weyl transforms of each subpatch.}
\label{fig:same_weyl}
\end{figure}

As an illustration of the above result, Figure~\ref{fig:same_weyl} displays a simple texture pattern, along with four patches taken from it at different positions and under different orientations. In the context of texture classification, it is desirable to have a representation which identifies all of these patches as coming from the same texture. In fact, all four patches can be obtained from each other by applying $D(a,b)$ matrices, and they therefore have the same Weyl transform coefficients in absolute value.

In Section~\ref{sec:intr}, we presented an example demonstrating the ability of the Weyl transform to effectively classify patches from two different textures (Fig.~\ref{fig:intro}). The Weyl transform distinguishes the two textures by capturing multiscale symmetries which are important in describing texture. Moreover, invariance to multiscale transformations (Theorem~\ref{thm:invariance}) ensures that the Weyl transforms of patches from the same texture exhibit similarity.

Each $D(a,b)$ matrix is a product of a permutation matrix and a diagonal `sign-change' matrix with $\pm 1$ entries. The sign-change patterns turn out to be columns of the Walsh-Hadamard transform matrix, also known as Walsh functions~\cite{hadamard_transform}. We will show in Section~\ref{sec:hadamard} that the Weyl transform can equally be viewed as a combination of autocorrelations and the Walsh-Hadamard transform, two familiar tools in signal processing (see Section~\ref{sec:intr}). This connection also provides a method for the fast computation of the Weyl transform by means of the fast (fully binary) Walsh-Hadamard transform (FWHT).

\subsection{The binary Heisenberg-Weyl group}
\label{sec:HWG}

The Weyl transform consists of inner products with signed permutation matrices from the binary Heisenberg-Weyl group, which we next describe in more detail. The Heisenberg-Weyl group can be thought of as a multiscale extension of the Dihedral group $D_8$, namely the symmetries of the square: four rotations through multiples of $90^{\circ}$, and reflections in its four axes of symmetry. These transformations can equally be viewed as matrix multiplications. Setting
\begin{equation}\label{eq:XZ_def}
X=\left(\begin{array}{cc}0&1\\1&0\end{array}\right)\;\;\mbox{and}\;\;Z=\left(\begin{array}{cc}1&0\\0&-1\end{array}\right),
\end{equation}
the matrix group $\left\{\pm X^a Z^b\;:\;a,b\in\ZZ_2\right\}$ provides a representation of $D_8$, where $\ZZ_2:=\{0,1\}$.

\begin{figure}[h!]
\centering
 {\includegraphics[angle=0, height=0.033\textwidth, width=.5\textwidth]{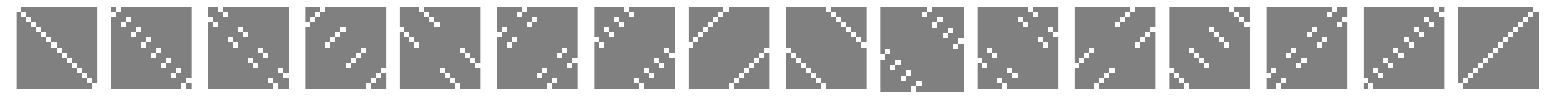} \hspace{0pt}}\\
 {\includegraphics[angle=0, height=0.033\textwidth, width=.5\textwidth]{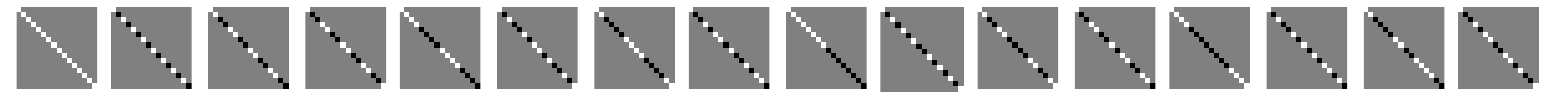} \hspace{0pt}}\\
{\includegraphics[angle=0, height=0.033\textwidth, width=.5\textwidth]{Dab.pdf}}
\caption{The signed permutation matrices $D(a,b)$ for $m=4$: -1 (black), 0 (grey), 1 (white). First row: Permutation matrices $D(a,0)$. Second row: Sign change matrices $D(0,b)$. Third row: Examples of matrices $D(a,b)$.}
\label{fig:dab}
\end{figure}

Next we use binary $m$-tuples to label the entries of a vector of length $2^m$ for some positive integer $m$. The $v$-th
coordinate is labeled by the binary expansion of $v$, and coordinates are ordered from right to left, so that $v=v_{m-1}2^{m-1}+\ldots +v_0$ is represented by $(v_{m-1}\ldots v_1\;v_0)^T\in\ZZ_2^m$ (where, for example, $\ZZ_2^2=\{(0\;0)^T,(0\;1)^T, (1\;0)^T, (1\;1)^T\}$). Given $a=(a_{m-1}\ldots a_0)^T\in\ZZ_2^m$, define $D(a,0)$ to be the permutation matrix given by the Kronecker product
$$D(a,0):=(X^{a_{m-1}})\otimes\ldots\otimes(X^{a_0}).$$
Note that the $D(a,0)$ are dyadic multiscale permutations, with the leftmost terms giving coarse-scale permutations and the rightmost terms giving fine-scale permutations. The first row of Fig.~\ref{fig:dab} displays the matrices $D(a,0)$ in the case of $m=4$. Similarly, given $b=(b_{m-1}\ldots b_0)^T\in\ZZ_2^m$, define $D(0,b)$ to be the sign change matrix given by the Kronecker product
$$D(0,b):=(Z^{b_{m-1}})\otimes\ldots\otimes(Z^{b_0}).$$
The matrices $D(0,b)$ for $m=4$ are displayed in the second row of Fig.~\ref{fig:dab}. We will show in Section~\ref{sec:hadamard} that the sign patterns are the columns of the Walsh-Hadamard transform matrix, also known as Walsh functions~\cite{hadamard_transform}. Define $D(a,b)$, for $a,b\in\ZZ_2^m$, by
\begin{equation}\label{eq:Dab_def}
D(a,b):=D(a,0)D(0,b),
\end{equation}
to obtain a collection of signed permutation matrices. The third row of Fig.~\ref{fig:dab} gives some examples of matrices $D(a,b)$ for $m=4$.

Define
\begin{equation}\label{eq:HW_def}
\mathcal{HW}_{2^m}:=\left\{i^\lambda D(a,b)\;:\;a,b\in\ZZ_2^m,\lambda\in\ZZ_4\right\}.
\end{equation}

The set $\mathcal{HW}_{2^m}$ forms the binary Heisenberg-Weyl group under matrix multiplication~\cite{finite_HW_radar}.

The set of all real symmetric matrices is a vector space, which we denote by $V$. Given real symmetric matrices $R,S\in V$, associate with $V$ the inner product $(R,S):=\mathrm{Tr}(R^T S)$, which induces the Frobenius norm $\|R\|_F:=[\mathrm{Tr}(R^T R)]^{\frac{1}{2}}$ on $V$. We now show that matrices $D(a,b)$ with $a^T b=0$ form a basis for $V$. Define 
$$\YY_2^m:=\left\{(a,b):a,b,\in\ZZ_2^m,a^T b=0\right\},$$ 
and 
$$\mathcal{B}_{2^m}:=\left\{\frac{1}{2^{m/2}}D(a,b):a,b\in\YY_2^m\right\}.$$
\vspace{4pt}
\begin{lemma}\label{lem:basis}
$\mathcal{B}_{2^m}$ is an orthonormal basis for $V$.
\end{lemma}

\textbf{Proof:}  If $a^T b=1$ then $D(a,b)$ squares to $-I$, and hence the inner product of $D(a,b)$ with any real symmetric matrix is zero. For more details see Section~\ref{invariance}.

Lemma~\ref{lem:basis} implies that any real symmetric matrix $R\in\RR^{2^m\times 2^m}$ can be expanded as
$$R=\sum_{(a,b)\in\YY_2^m}\left\{\frac{1}{2^{m/2}}\mathrm{Tr}[R\cdot D(a,b)]\right\}\frac{1}{2^{m/2}}D(a,b).$$
In particular, given a vectorized signal $y\in\RR^{2^m}$, its correlation matrix $yy^T$ can be expanded as
\begin{eqnarray}\label{eq:expand}
yy^T&=&\sum_{(a,b)\in\YY_2^m}\left\{\frac{1}{2^{m/2}}\mathrm{Tr}[yy^T\cdot D(a,b)]\right\}\frac{1}{2^{m/2}}D(a,b)\nonumber\\
&=&\sum_{(a,b)\in\YY_2^m}\omega_{a,b}(y)\frac{1}{2^{m/2}}D(a,b).
\end{eqnarray}
The coefficients $\omega_{a,b}(y)$ are the \emph{Weyl coefficients} of $y$ as introduced in Section~\ref{sec:summary}. When we connect the Weyl transform to autocorrelation in Section~\ref{sec:hadamard}, it will be convenient to pad with zeros by defining $\omega_{a,b}:=0$ when $a^T b=1$. Since we have expanded in an orthonormal basis, the mapping from $yy^T$ to $\left\{\omega_{a,b}\right\}$ is an isometry, namely the \emph{Weyl transform} as introduced in Section~\ref{sec:summary}.

The Weyl transform also has a striking geometrical interpretation. Each $D(a,b)$ matrix (except for the identity) has $\pm 1$ eigenspaces of multiplicity $m/2$ respectively, so that
$$\begin{array}{rcl}D(a,b)&=&\begin{bmatrix}P_{a,b}&Q_{a,b}\end{bmatrix}\begin{bmatrix}I&0\\0&-I\end{bmatrix}\begin{bmatrix}P_{a,b}^T\\Q_{a,b}^T\end{bmatrix}\\
&=&P_{a,b}P_{a,b}^T-Q_{a,b}Q_{a,b}^T,
\end{array}$$
where $P_{a,b}$ and $Q_{a,b}$ are orthonormal bases for the $+1$ and $-1$ eigenspaces respectively. The Weyl coefficient $\omega_{a,b}$ can therefore be expressed as
$$\begin{array}{rcl}\omega_{a,b}&=&\frac{1}{2^{m/2}}\mathrm{Tr}[yy^T(P_{a,b}P_{a,b}^T-Q_{a,b}Q_{a,b}^T)]\\
&=&\frac{1}{2^{m/2}}\left\{\|P_{a,b}^Ty\|_F^2-\|Q_{a,b}^Ty\|_F^2\right\},
\end{array}$$
which reveals that $\omega_{a,b}$ gives information about the relative distance of the covariance matrix $yy^T$ from two half-spaces. A large positive value means that the image is in the $+1$ eigenspace; a large negative value means that the image is in the $-1$ eigenspace. A large absolute value of either sign indicates that the image exhibits periodic symmetry.

It is shown in~\cite{group_theoretic} that the collection of half-spaces induced by the Weyl transform is in fact the optimal half-space packing originally given in~\cite{shor_sloane}\footnote{It is also shown in~\cite{shor_sloane} that this subspace packing can be constructed recursively.}. The Weyl transform can therefore be viewed as a principled matched filter for covariance matrices.

\subsection{The Weyl transform and autocorrelation: a connection}\label{sec:hadamard}

We next establish the bridge between Weyl coefficients and autocorrelation, showing that the Weyl transform can equally be viewed in terms of the Walsh-Hadamard transform~\cite{hadamard_transform} of binary autocorrelations. Consider a vectorized signal $y\in\RR^{2^m}$ indexed by a binary $m$-tuple $v=(v_{m-1}\ldots v_0)^T$. Then the correlation matrix $yy^T$ may be divided into $2^m$ autocorrelation bands $z_a\in\RR^{2^m}$, where 
\begin{equation}\label{bands}
(z_a)_v:=y_v y_{v+a}.
\end{equation} 
Each autocorrelation band $z_a$ gives information on the invariance of a signal to a particular binary translation $v\rightarrow v+a$.
Given a binary $m$-tuple $a$, define $\omega_a\in\RR^{2^m}$ to be the subset of Weyl coefficients indexed by $a$, that is $(\omega_a)_b:=\omega_{a,b}$. Define the $2^m\times 2^m$ Walsh-Hadamard transform matrix $H_{2^m}$ by 
$$(H_{2^m})_{v,w}:=\frac{1}{2^{m/2}}(-1)^{v^T w},$$ 
where $v$ and $w$ are binary $m$-tuples~\cite{hadamard_transform}. The following result characterizes the Weyl transform as a combination of autocorrelations and the Walsh-Hadamard transform.\vspace{3pt}

\begin{theorem}[\textbf{Weyl transform in terms of autocorrelation}]\label{thm:hadamard}
Let the autocorrelation bands of $y$, $\{z_a\}$, be defined as in (\ref{bands}). Then
$\omega_a(y)=H_{2^m}z_a$,
where $\{\omega_a\}_b=0$ if $a^T b=1$.
\end{theorem}
\textbf{Proof:} See Section~\ref{app:autocorrelation}.

\section{Illustrative examples: The invariance and versatility of the Weyl transform}
\label{sec:exa}

We next further illustrate fundamental properties of the Weyl transform using real-world texture examples.
We suggest here two ways to exploit the Weyl transform for effective signal representation: equivalence class histograms and supervised coefficient selection. We first exploit the underlying group structure of the transform to build invariance to particular geometrically-significant transformations; and then we describe how training data can be used to select the Weyl coefficients that are most significant in distinguishing the classes.

\begin{figure*} [ht]
\centering
 \subfloat[\emph{Jeans}] {\includegraphics[angle=0, height=0.11\textwidth, width=.11\textwidth]{jpg/bluejeans.jpg} \hspace{0pt}}
  \subfloat[\emph{CottonB}] {\includegraphics[angle=0, height=0.11\textwidth, width=.11\textwidth]{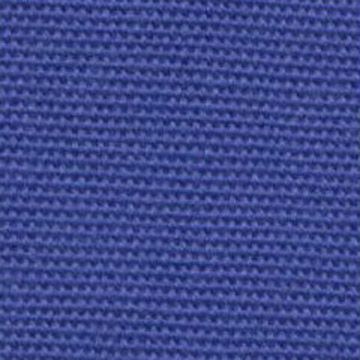} \hspace{0pt}}
    \subfloat[\emph{CottonG}] {\includegraphics[angle=0, height=0.11\textwidth, width=.11\textwidth]{jpg/greenCotton.jpg} \hspace{0pt}}
  \subfloat[\emph{FabricG}] {\includegraphics[angle=0, height=0.11\textwidth, width=.11\textwidth]{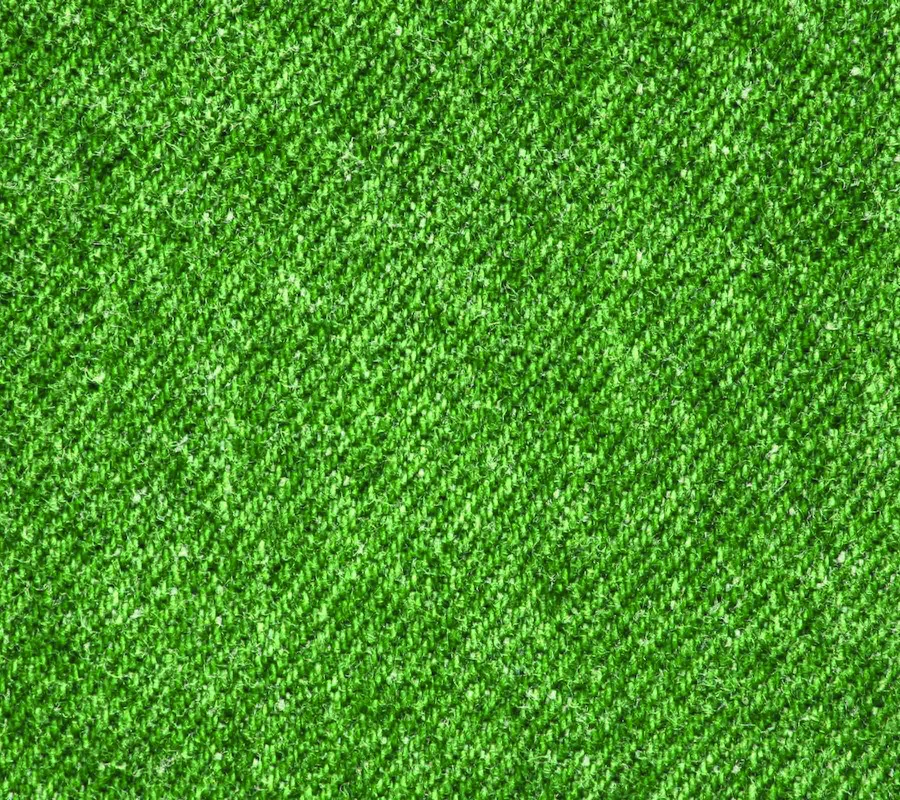} \hspace{0pt}}
    \subfloat[\emph{Fabric}] {\includegraphics[angle=0, height=0.11\textwidth, width=.11\textwidth]{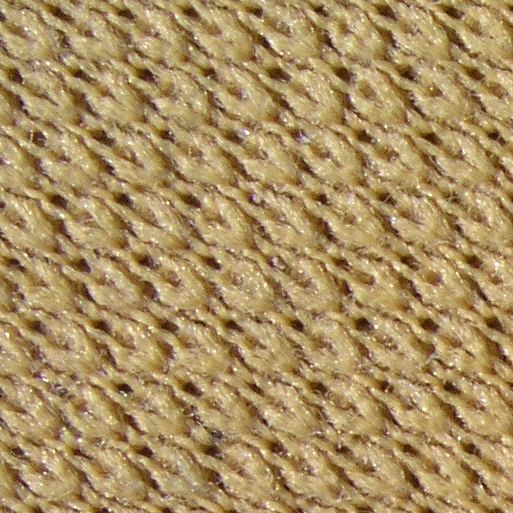} \hspace{0pt}}
      \subfloat[\emph{TextileG}] {\includegraphics[angle=0, height=0.11\textwidth, width=.11\textwidth]{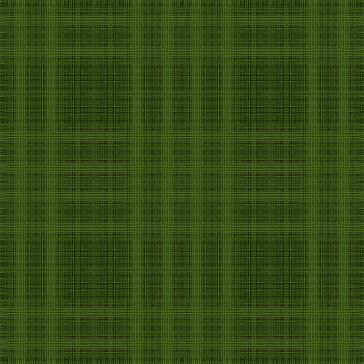} \hspace{0pt}}
        \subfloat[\emph{TextileB}] {\includegraphics[angle=0, height=0.11\textwidth, width=.11\textwidth]{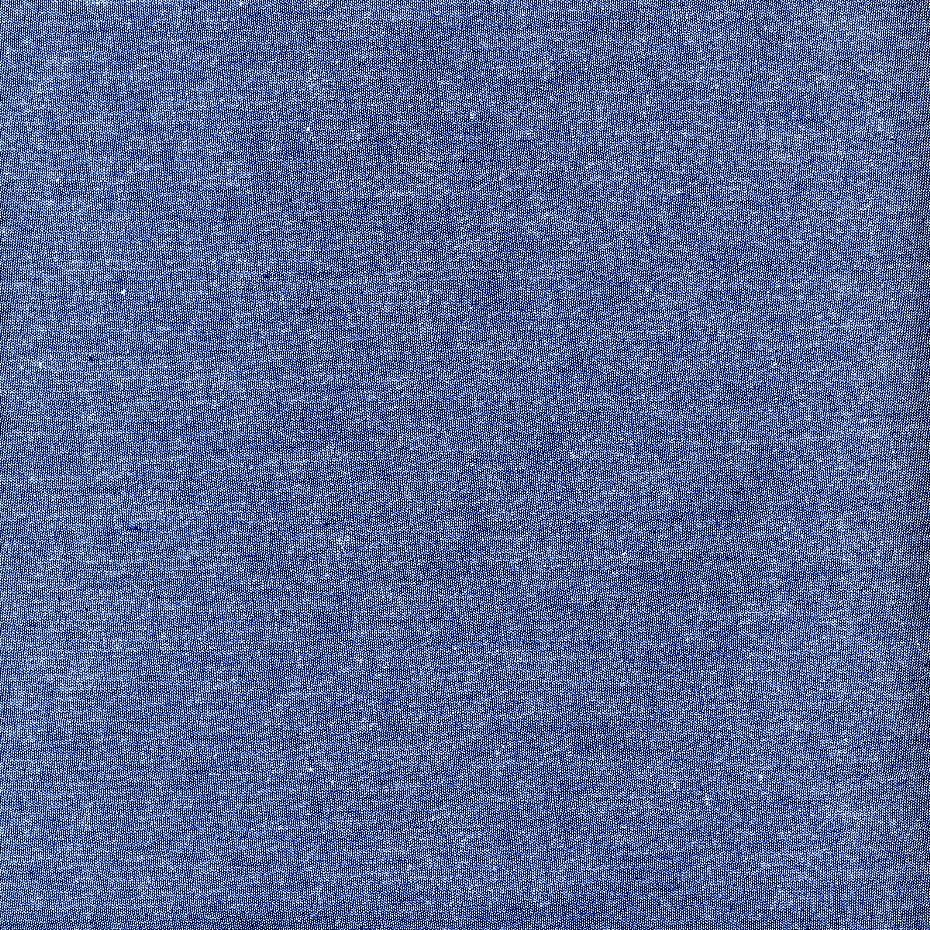}}
\caption{Seven fabric textures.}
\label{fig:fabric}
\end{figure*}

\subsection{Equivalence class histograms}

While the absolute values of the Weyl coefficients are already invariant under Heisenberg-Weyl transformations, more can be said if we allow the coefficients to be permuted (but unchanged in absolute value). Given an arbitrary linear transformation $y\rightarrow \phi y$, the cyclic property of the trace implies that
$$\begin{array}{rcl}
\omega_{a,b}(\phi y)&=&\frac{1}{2^{m/2}}\mathrm{Tr}[\phi yy^T\phi^T D(a,b)]\\
&=&\frac{1}{2^{m/2}}\mathrm{Tr}[yy^T\phi^T D(a,b)\phi],
\end{array}$$
which in turn implies that $\phi$ permutes the Weyl coefficient absolute values if it is an outer automorphism of $\mathcal{HW}_{2^m}$. The outer automorphism group of the Heisenberg-Weyl group is known: it is the binary symplectic group $\mathrm{Sp}(2m,\ZZ_2)$~\cite{finite_HW_radar}. Based on these observations, we may conceive of a method of pooling Weyl coefficients which builds invariance to any desired symplectic transformation, which we next describe.
While often, e.g., in deep learning, pooling is heuristically designed \cite{pooling}, here we provide a principled approach to it \cite{Bruna, scattering}.

Let $G$ be some (sub)group of symplectic transformations. Then $G$ acts on $\mathcal{HW}_{2^m}$ by conjugation, and the action partitions $\mathcal{HW}_{2^m}$ into equivalence classes. This means that the group $G$ can only permute the elements of $\mathcal{HW}_{2^m}$ within each equivalence class. It follows that the average of the absolute values of the Weyl coefficients within a given equivalence class is invariant under $G$. Such equivalence class averaging has the further appeal of reducing the number of Weyl coefficients, and it is also extremely versatile  since any group of geometrically-significant symplectic transformations can be considered.

As an example, we illustrate the proposed approach in the context of a vectorized $N\times N$ image $y$, taking $G$ to be the group of transformations generated by $90^{\circ}$ rotation and cyclic horizontal and vertical translation by any multiple of $N/4$. We first show that these are each indeed symplectic transformations. We assume that our original image $Y$ has dimensions $N\times N$, $N=2^r$, and that it is vectorized columnwise to give $y$, a vector of length $2^m$ where $m=2r$. Let $a,b\in\ZZ_2^{2r}$, and let us write $a=(a_{2r-1}\ldots a_0)^T$ and $b=(b_{2r-1}\ldots b_0)^T$. We write $1_r$ for the vector of $r$ ones.
\vspace{4pt}
\begin{proposition}[\textbf{Symplectic permutations}]\label{lem:symplectic}
(i) Rotation of $Y$ by $90^{\circ}$ clockwise corresponds to the mapping
\begin{equation}\label{eq:rotation}
D\left(\left[\begin{array}{c}a_1\\a_2\end{array}\right],\left[\begin{array}{c}b_1\\b_2\end{array}\right]\right)\longrightarrow (-1)^{1^T b_1}D\left(\left[\begin{array}{c}a_2\\a_1\end{array}\right],\left[\begin{array}{c}b_2\\b_1\end{array}\right]\right),
\end{equation}
where $a_1,a_2,b_1,b_2\in\ZZ_2^r$.\\
(ii) Cyclic translation of $Y$ by $N/4$ vertically corresponds to the mapping
\begin{equation}\label{eq:translation}
D\left(\left[\begin{smallmatrix}a_1\\j\\k\\a_2\end{smallmatrix}\right],\left[\begin{smallmatrix}b_1\\l\\m\\b_2\end{smallmatrix}\right]\right)\longrightarrow (-1)^{m}D\left(\left[\begin{smallmatrix}a_1\\j+k\\k\\a_2\end{smallmatrix}\right],\left[\begin{smallmatrix}b_1\\l\\l+m\\b_2\end{smallmatrix}\right]\right),
\end{equation}
where $a_1,b_1\in\ZZ_2^r$, $a_2,b_2\in\ZZ_2^{r-2}$ and $j,k,l,m\in\ZZ_2$.
\end{proposition}

\textbf{Proof:} See Section~\ref{app:symplectic}.

\begin{figure*} [ht]
\centering
 \subfloat[Weyl (\textbf{86.49}\%)] {\includegraphics[angle=0, height=0.16\textwidth, width=.2\textwidth]{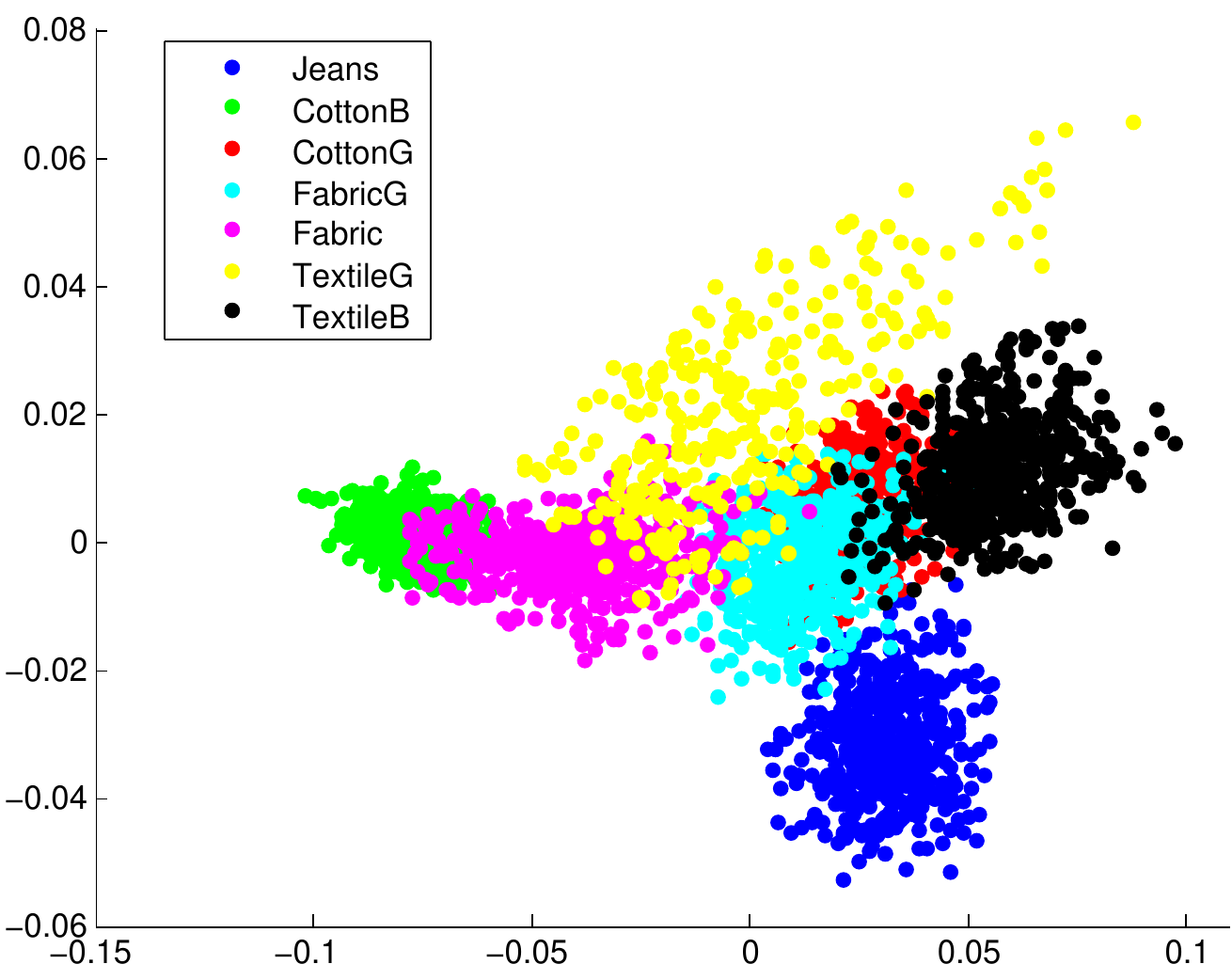}}
  \subfloat[Intensity (50.82\%)] {\includegraphics[angle=0, height=0.16\textwidth, width=.2\textwidth]{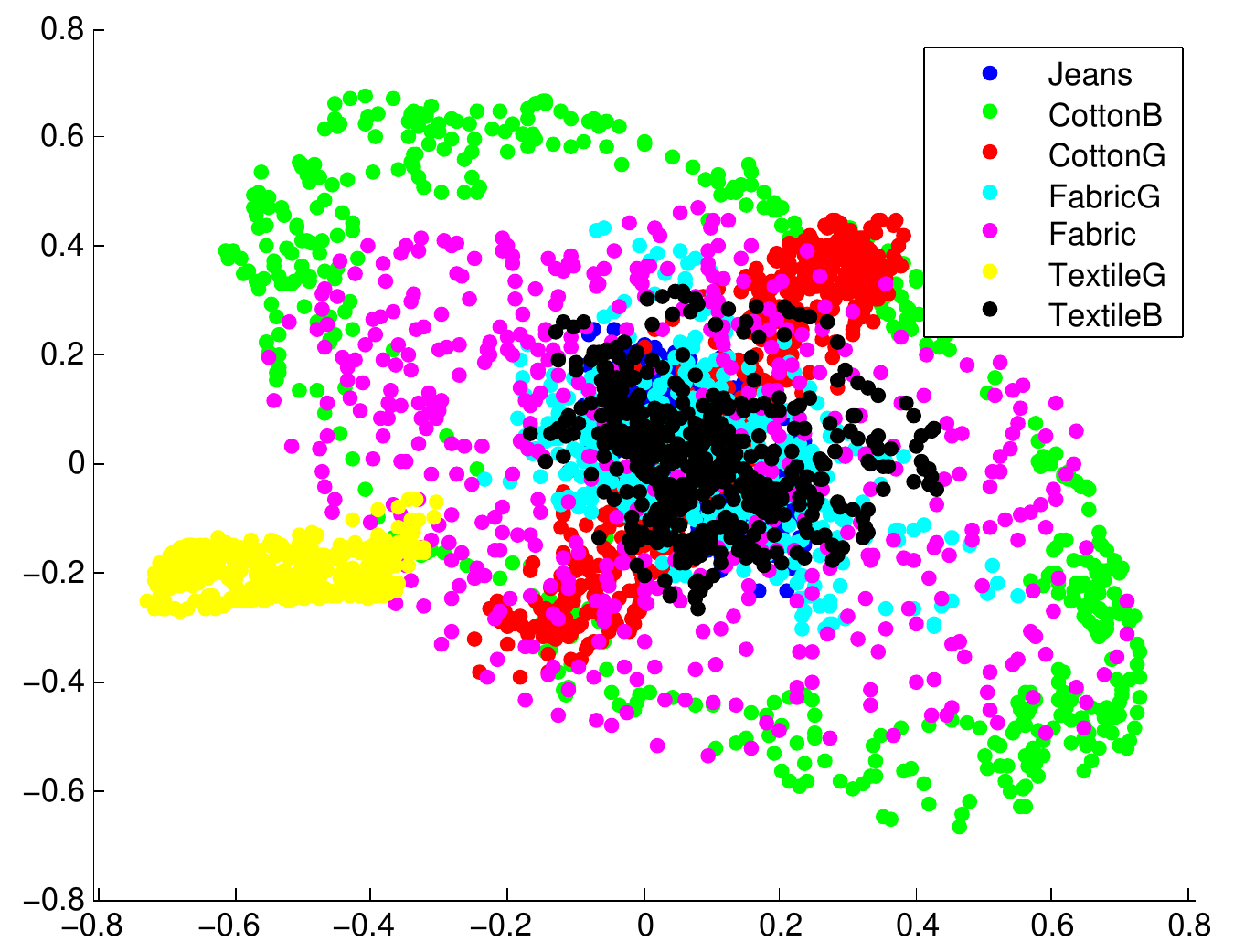}}
    \subfloat[HOG (78.28\%)] {\includegraphics[angle=0, height=0.16\textwidth, width=.2\textwidth]{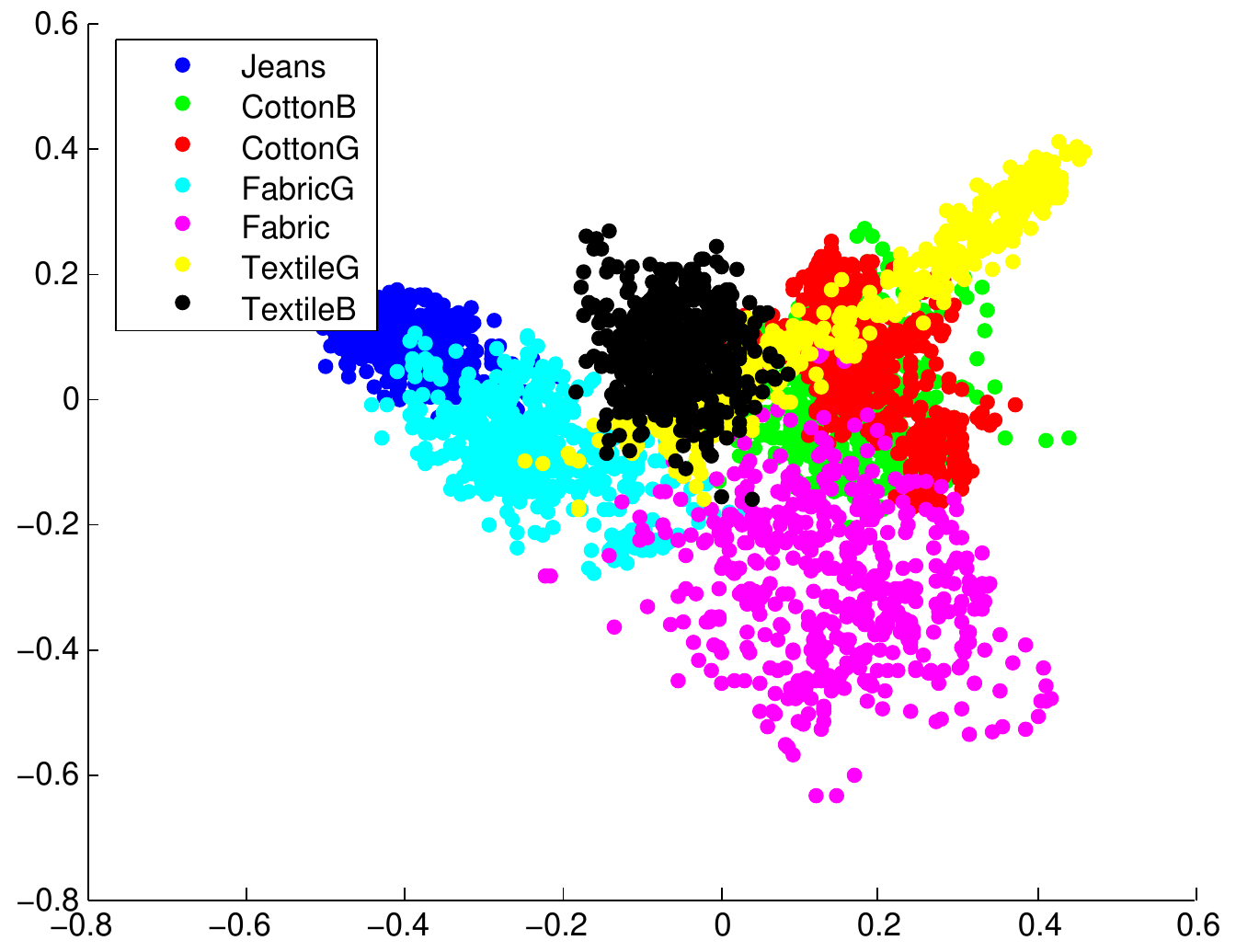}}
  \subfloat[Gabor (48.48\%)] {\includegraphics[angle=0, height=0.16\textwidth, width=.2\textwidth]{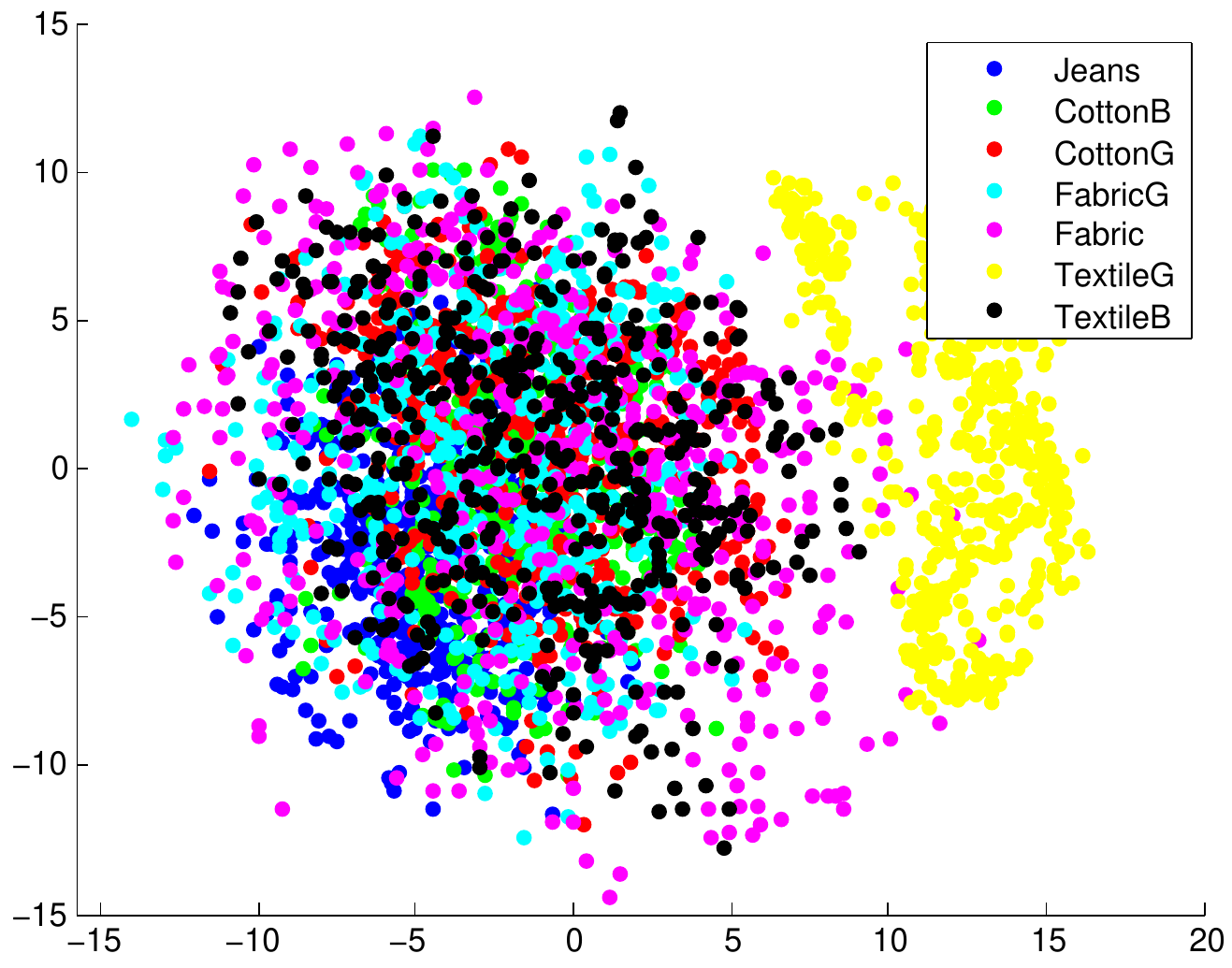}}
    \subfloat[LBP (78.71\%)] {\includegraphics[angle=0, height=0.16\textwidth, width=.2\textwidth]{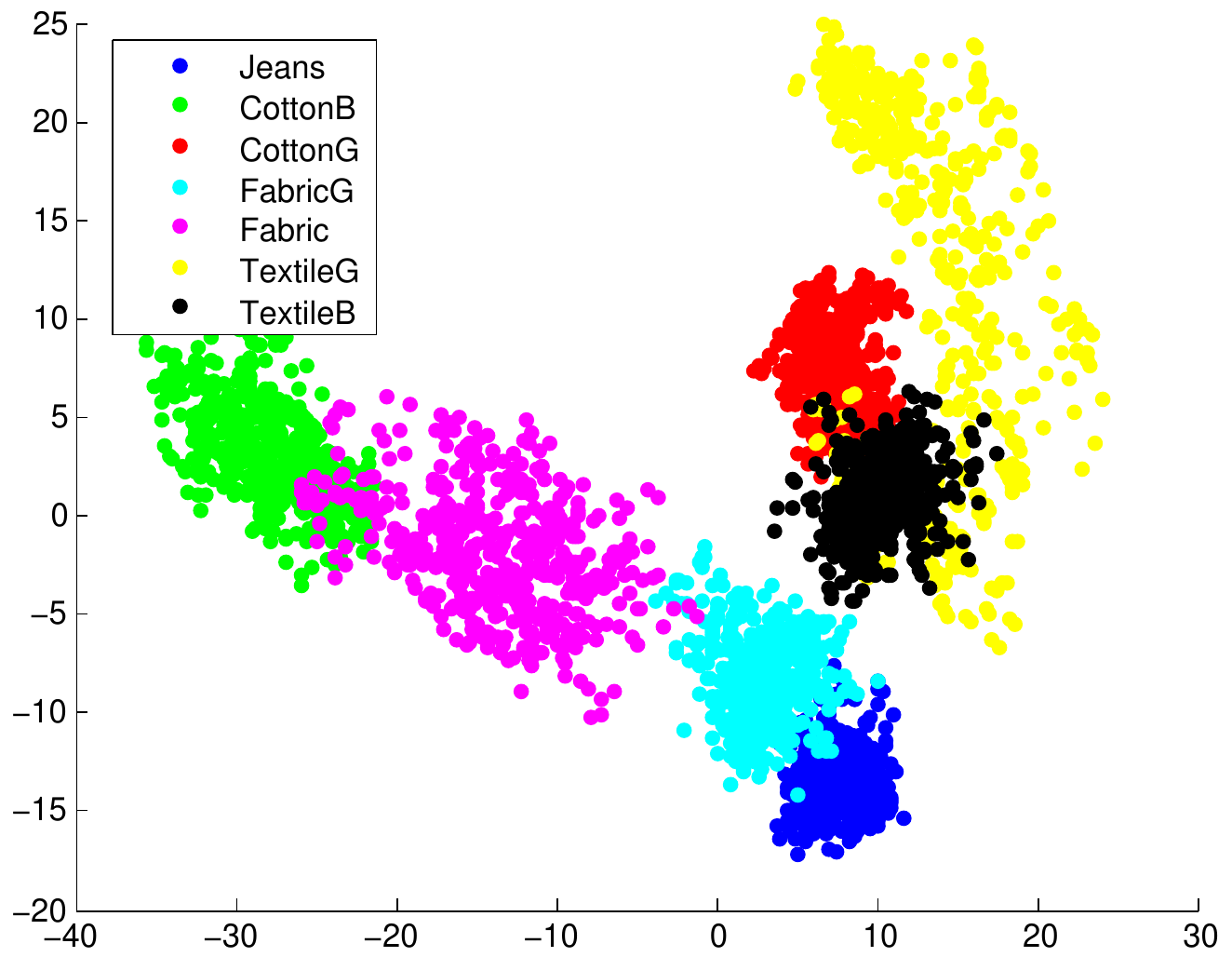}}
\caption{Fabric texture patches represented with different descriptors. Data are visualized with the dimension reduced to 2 using PCA. Different textures are shown in different colors:
\emph{Jeans} (blue), \emph{CottonB} (green), \emph{CottonG} (red), \emph{FabricG} (cyan), \emph{Fabric} (magenta), \emph{TextileG} (yellow), \emph{TextileB} (black).
The k-means clustering accuracies in  parentheses approximately assess the discriminability (and class compactness) of each descriptor. The respective size of each descriptor is  Weyl (24), Intensity (256), HOG (576), Gabor (640) and LBP (256).
The proposed Weyl descriptor is both the most compact and discriminative.
}
\label{fig:fabricRep}
\end{figure*}

\begin{figure*} [ht]
\centering
 \subfloat[Weyl (\textbf{84.62}\%)] {\includegraphics[angle=0, height=0.16\textwidth, width=.2\textwidth]{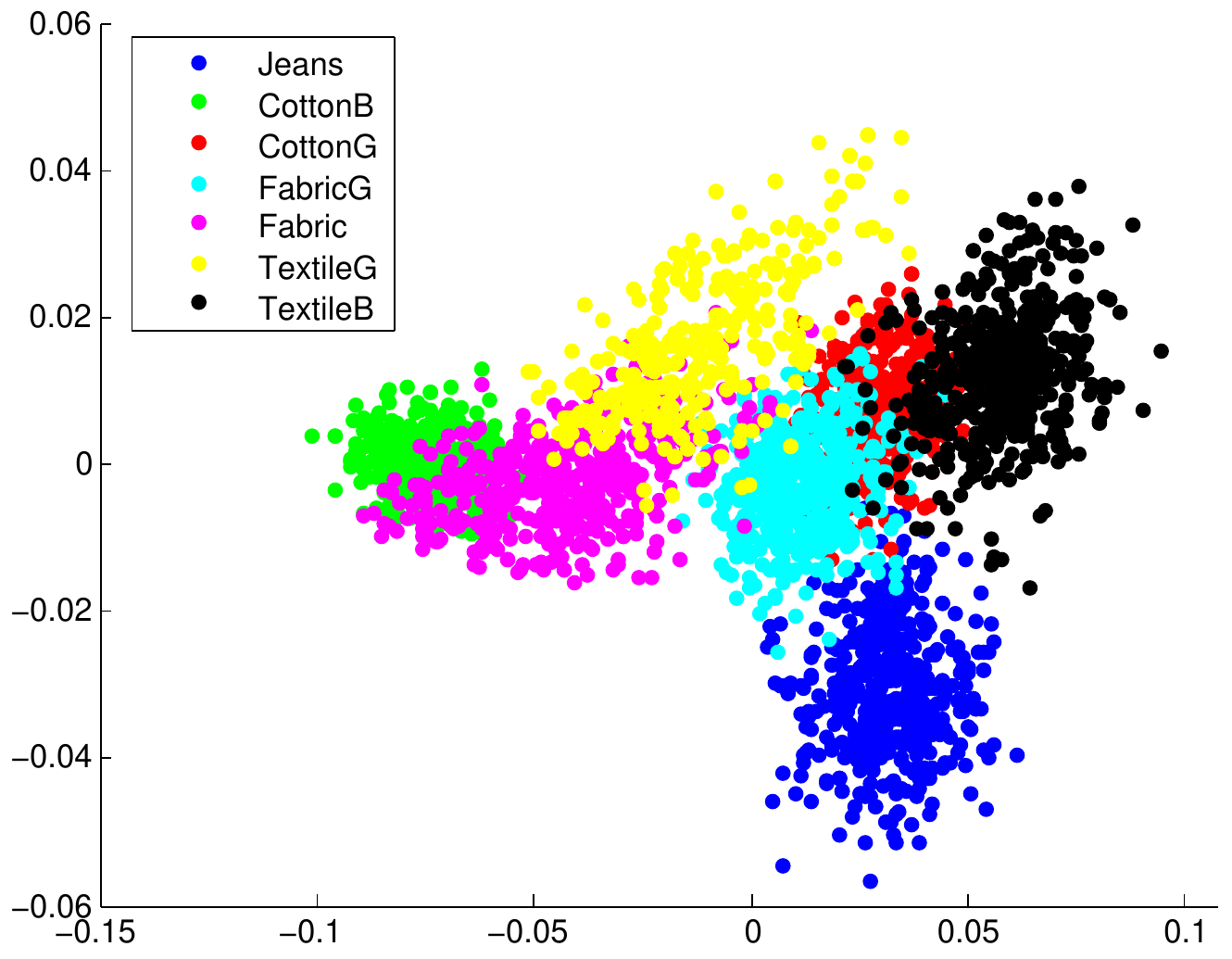} \hspace{0pt}}
  \subfloat[Intensity (33.65\%)] {\includegraphics[angle=0, height=0.16\textwidth, width=.2\textwidth]{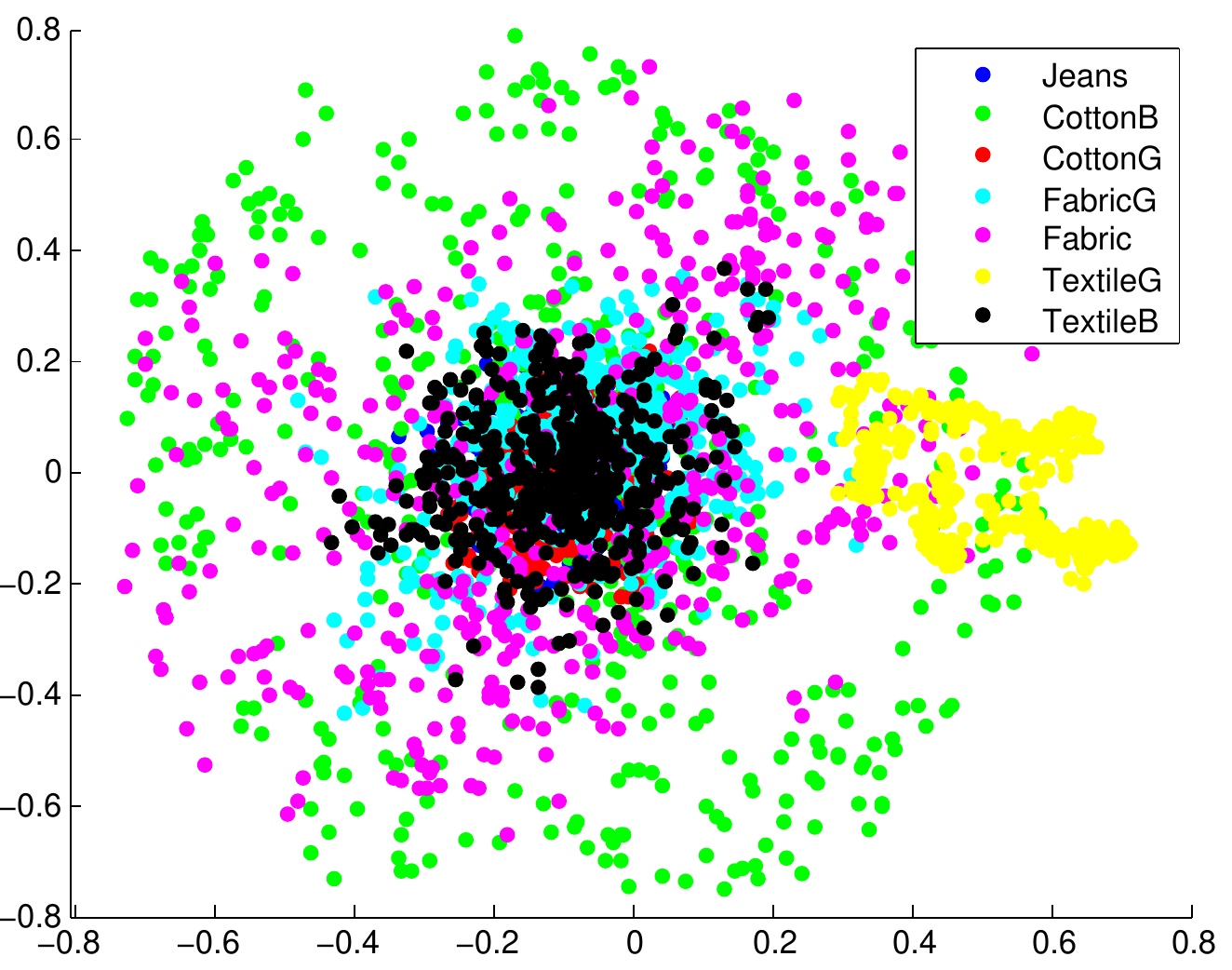}}
    \subfloat[HOG (30.00\%)] {\includegraphics[angle=0, height=0.16\textwidth, width=.2\textwidth]{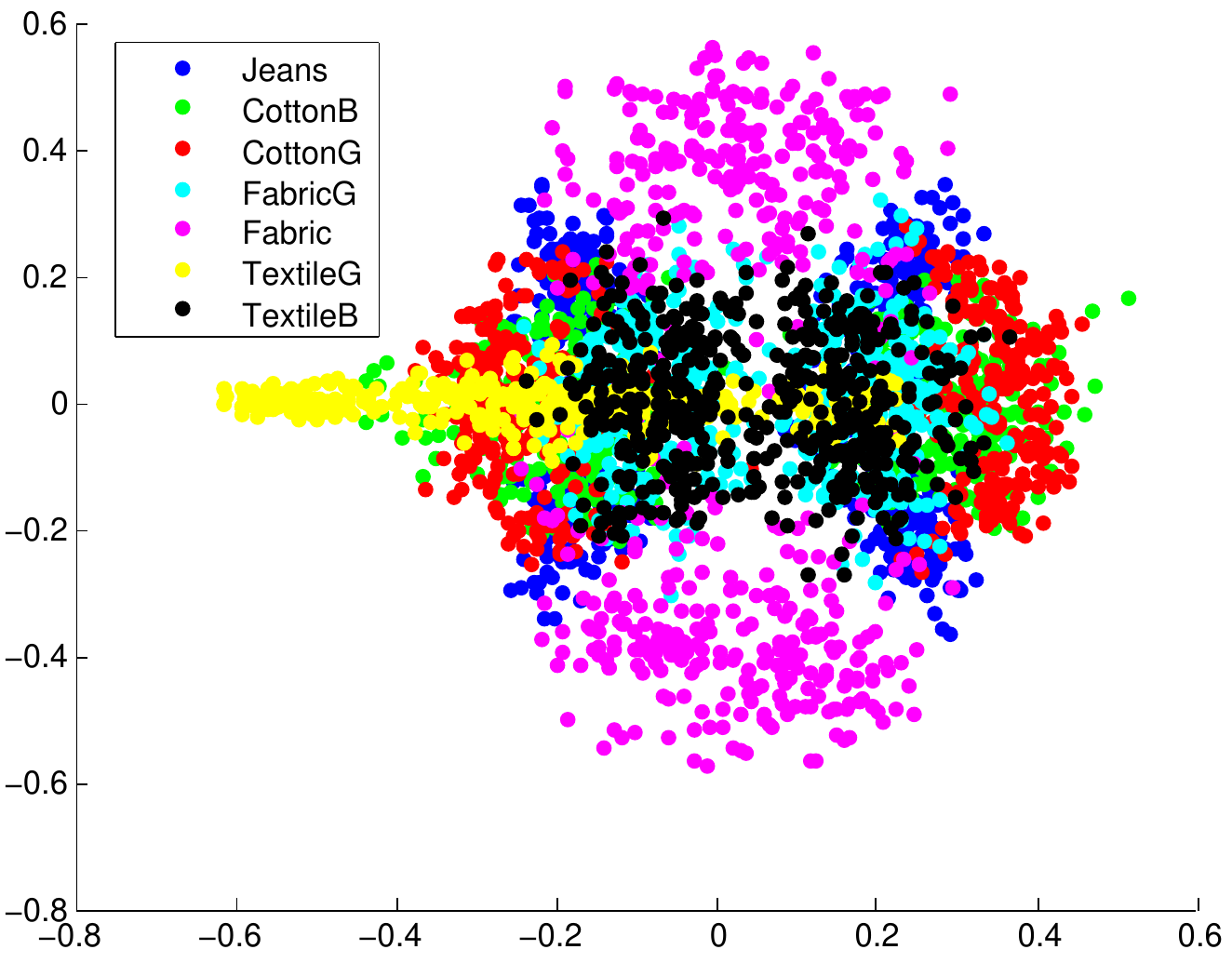}}
  \subfloat[Gabor (32.65\%)] {\includegraphics[angle=0, height=0.16\textwidth, width=.2\textwidth]{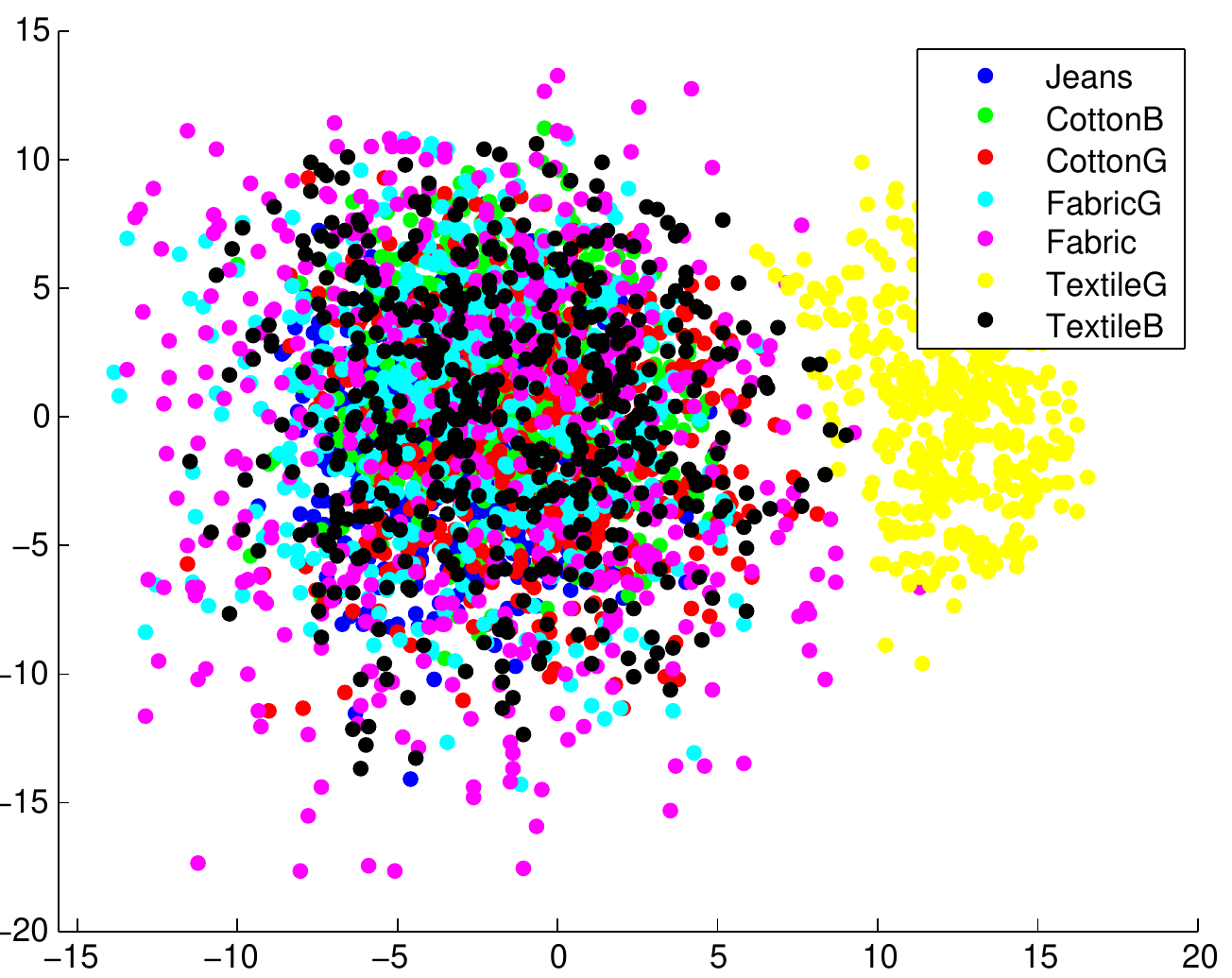}}
    \subfloat[LBP (53.40\%)] {\includegraphics[angle=0, height=0.16\textwidth, width=.2\textwidth]{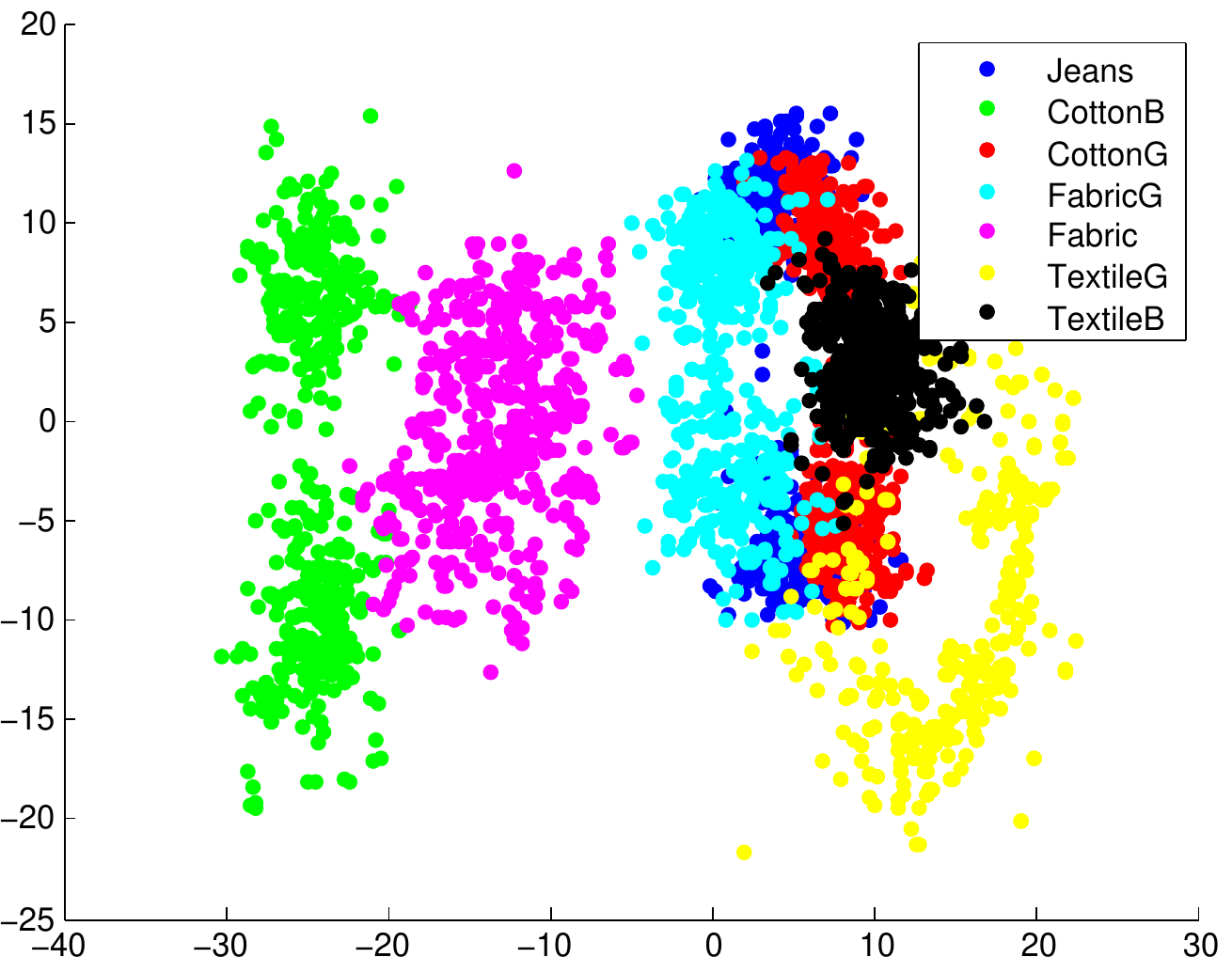}}
\caption{Fabric texture patches under the dihedral transformations represented with different descriptors (see Fig.~\ref{fig:fabricRep} for notation description).
Comparing to Fig.~\ref{fig:fabricRep}, we observe that the Weyl representation is invariant under dihedral group transformations.  For other descriptors, we observe new sub-clusters, indicating that significantly different descriptors are assigned to the same texture under dihedral transformations.
}
\label{fig:fabricRepDih}
\end{figure*}

An analogous result for cyclic translation by $N/4$ horizontally can be obtained by permuting the vertical and horizontal coordinates. Note that a cyclic translation by $N/2$ fixes the Weyl coefficients' absolute values, revealing the transformation to be an element of the Heisenberg-Weyl group.

For illustration, in the following experiments, we consider the case of a $4\times 4$ image patch, vectorized to give $y\in\RR^{16}$, so that its Weyl coefficients $\omega_{a,b}$ are indexed by a pair of $4$-tuples. Let $G$ be the group of transformations generated by $90^{\circ}$ rotation and cyclic translation by $N/4$ (in this case, all possible cyclic translations) either vertically or horizontally. $G$ partitions the $a$ coordinates into the $6$ equivalence classes
\begin{equation}\label{eq:a_classes}
\begin{array}{c}
\big\{(0\;0\;0\;0)\big\},\;\big\{(1\;0\;1\;0)\big\},\;\big\{(0\;0\;1\;0),(1\;0\;0\;0)\big\},\\
\big\{(0\;0\;0\;1),(0\;0\;1\;1),(0\;1\;0\;0),(1\;1\;0\;0)\big\},\\
\big\{(0\;1\;0\;1),(0\;1\;1\;1),(1\;1\;0\;1),(1\;1\;1\;1)\big\},\\
\big\{(0\;1\;1\;0),(1\;0\;0\;1),(1\;0\;1\;1),(1\;1\;1\;0)\big\},
\end{array}
\end{equation}
while, independently, $G$ also partitions the $b$ coordinates into the $6$ equivalence classes
\begin{equation}\label{eq:b_classes}
\begin{array}{c}
\big\{(0\;0\;0\;0)\big\},\;\big\{(0\;1\;0\;1)\big\},\;\big\{(0\;0\;0\;1),(0\;1\;0\;0)\big\},\\
\big\{(0\;0\;1\;0),(0\;0\;1\;1),(1\;0\;0\;0),(1\;1\;0\;0)\big\}\\
\big\{(1\;0\;1\;0),(1\;0\;1\;1),(1\;1\;1\;0),(1\;1\;1\;1)\big\},\\
\big\{(1\;0\;0\;1),(0\;1\;1\;0),(0\;1\;1\;1),(1\;1\;0\;1)\big\}.
\end{array}
\end{equation}
We thus partition the Weyl coefficients into $36$ equivalence classes.

Six of the classes contain Weyl coefficients which are all zero, since $a^T b=1$ for all members of the class, and so we discard these classes. In addition, the six classes for which $a=0$ give variance information rather than autocorrelation information, so we also discard these classes. We are left with $24$ classes, within each of which we average the Weyl coefficient absolute values. Since $G$ permutes the members of each class, these averages are invariant to all transformations in $G$. We have thereby reduced the number of Weyl coefficients from $136$ to $24$, creating a representation that is invariant under not only the Heisenberg-Weyl group, but also the group $G$. In particular, compactness means that different ways of orienting and translating the same texture produce the same descriptor. Note that this equivalent class histogram approach is versatile: any geometrically-significant symplectic transformations could be quotiented out in this way depending on the expected image symmetries to be encountered.

\textbf{Fabric texture examples.} Seven real-world fabric textures are shown in Fig.~\ref{fig:fabric}. In this set of experiments, we randomly sample 500 $16\times16$-sized gray-scale patches from each texture, and mix them up.
The obtained 3500 texture patches are represented using 5 different descriptors: Weyl, intensity (the gray-scale value), Gabor \cite{Gabor}, LBP \cite{LBP}, and HOG \cite{HOG}.
With the equivalence classes defined above, 24 Weyl coefficients (absolute values) are required to represent each $4\times4$ patch.
For a $16\times16$ patch, we further average across the 16 obtained 24-sized vectors to form a 24-sized Weyl descriptor.
The Gabor features used here have 5 scales and 8 orientations, down-sampled by a factor of 4, and each feature vector is of length 640 .
We use the basic LBP histogram of length 256.
 For HOG features, we use a block size of $4 \times 4$ and 36 histogram bins, which gives a 576-sized feature vector. The proposed Weyl representation is therefore the most compact one.

\begin{figure*} [t]
\centering
 \subfloat[\emph{Sand}] {\includegraphics[angle=0, height=0.11\textwidth, width=.11\textwidth]{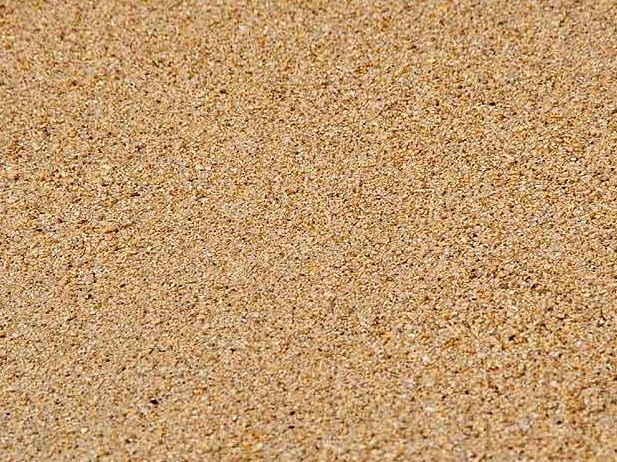} \hspace{0pt}}
  \subfloat[\emph{Rock}] {\includegraphics[angle=0, height=0.11\textwidth, width=.11\textwidth]{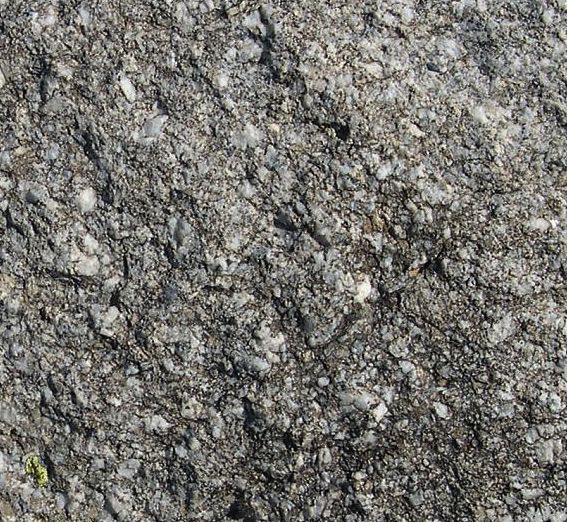} \hspace{0pt}}
    \subfloat[\emph{Sea}] {\includegraphics[angle=0, height=0.11\textwidth, width=.11\textwidth]{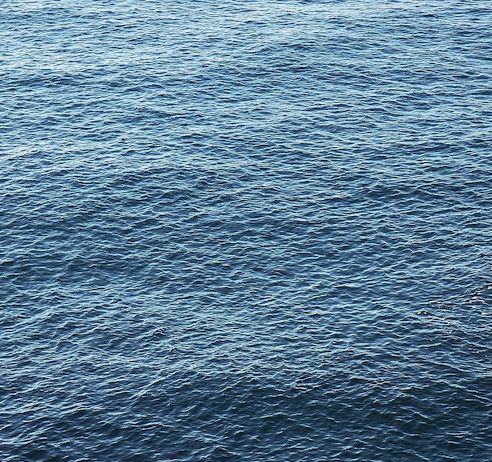} \hspace{0pt}}
  \subfloat[\emph{Sky}] {\includegraphics[angle=0, height=0.11\textwidth, width=.11\textwidth]{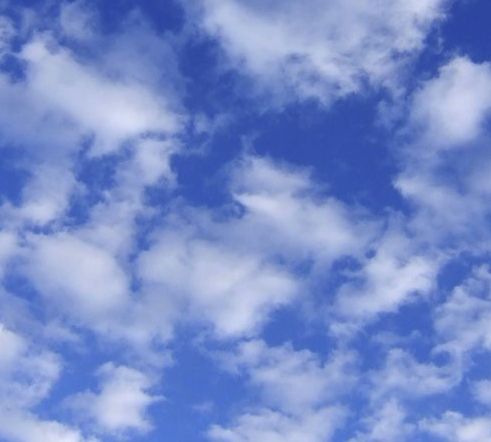} \hspace{0pt}}
    \subfloat[\emph{Wood}] {\includegraphics[angle=0, height=0.11\textwidth, width=.11\textwidth]{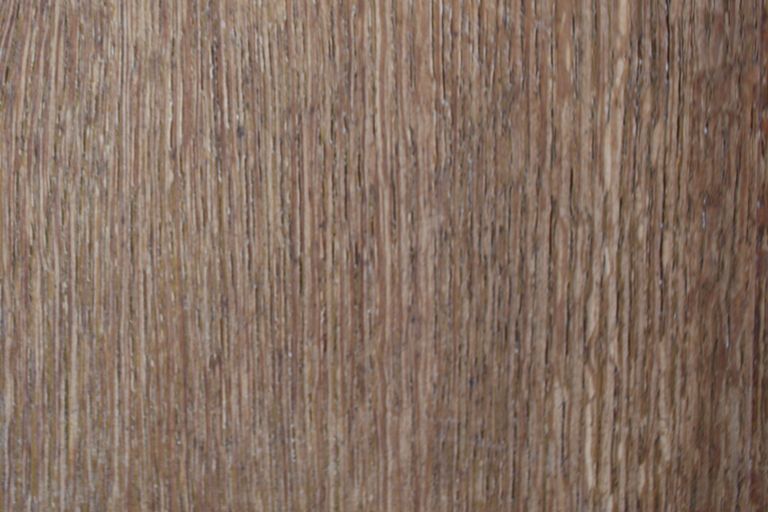}}
\caption{Five nature textures.}
\label{fig:nature}
\end{figure*}

\begin{figure*} [t]
\centering
 \subfloat[Weyl (\textbf{70.78}\%)] {\includegraphics[angle=0, height=0.16\textwidth, width=.2\textwidth]{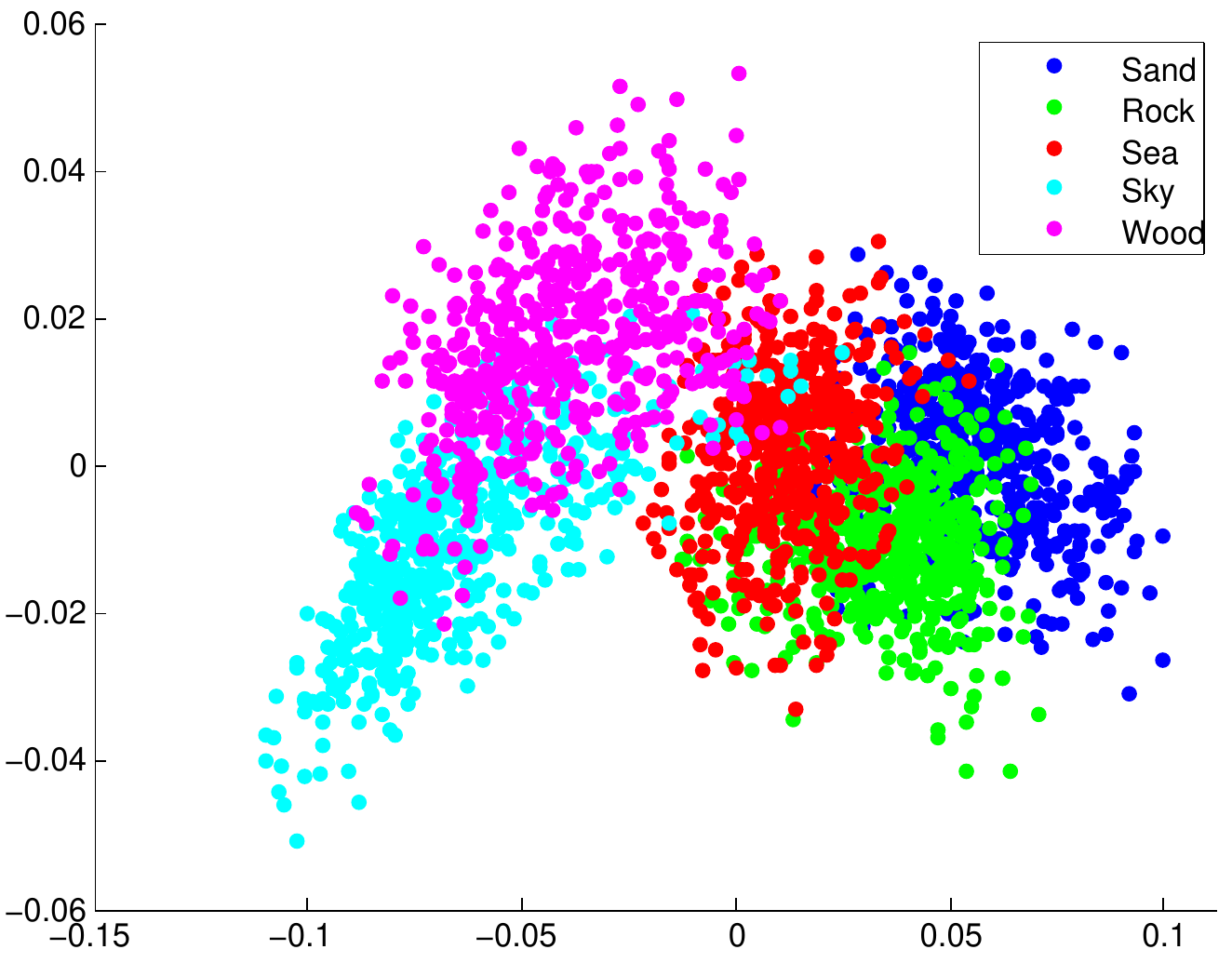} \hspace{0pt}}
  \subfloat[Intensity (40.16\%)] {\includegraphics[angle=0, height=0.16\textwidth, width=.2\textwidth]{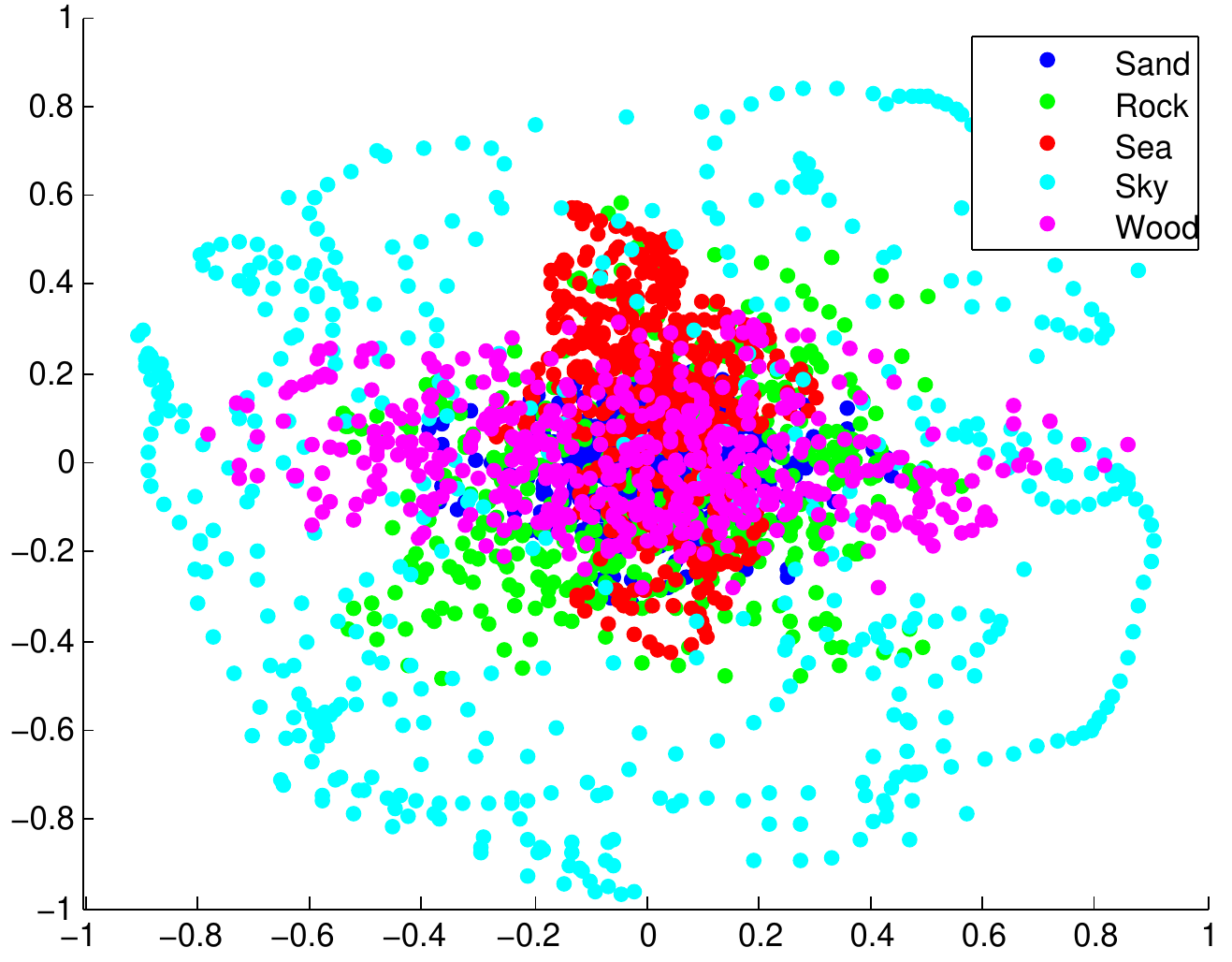}}
    \subfloat[HOG (61.11\%)] {\includegraphics[angle=0, height=0.16\textwidth, width=.2\textwidth]{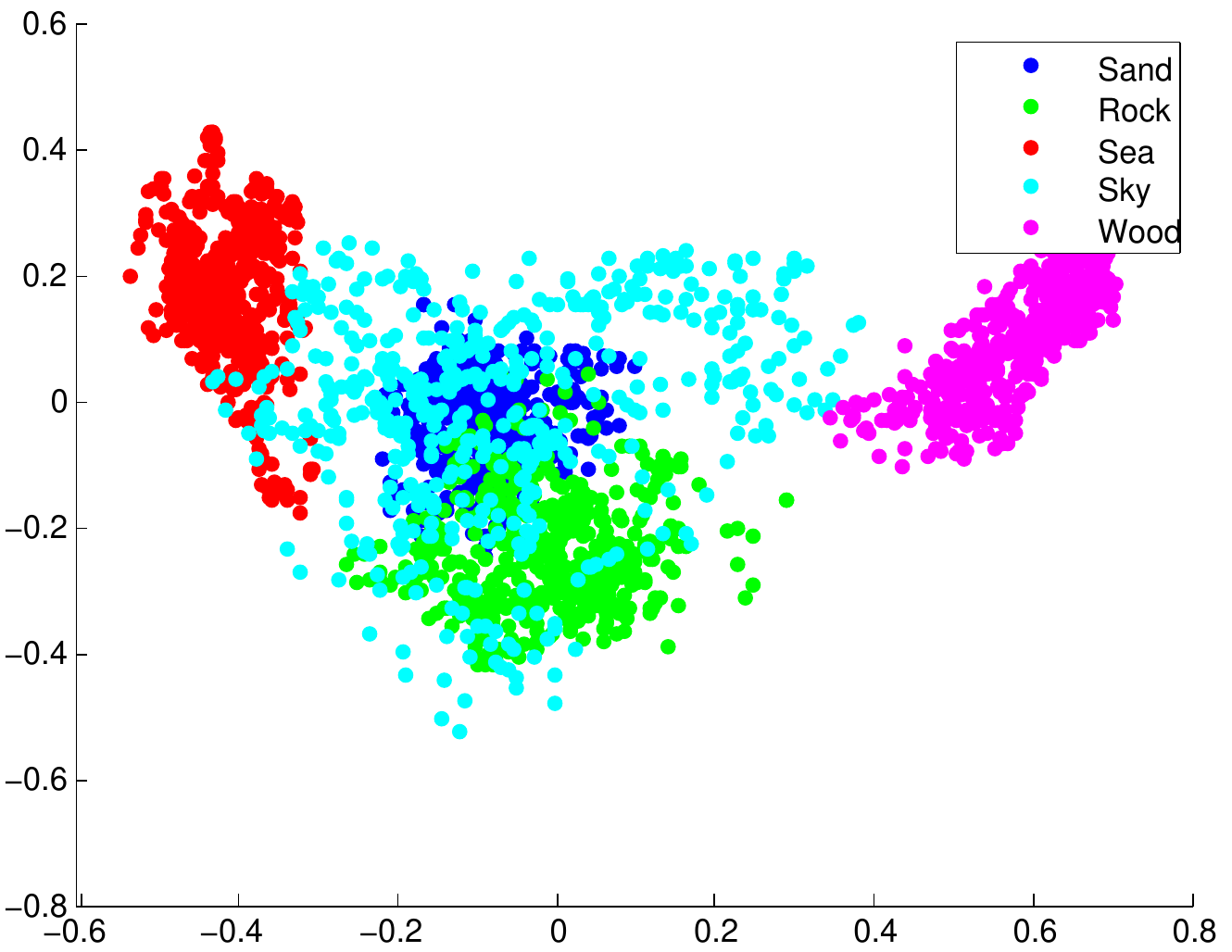}}
  \subfloat[Gabor (26.48\%)] {\includegraphics[angle=0, height=0.16\textwidth, width=.2\textwidth]{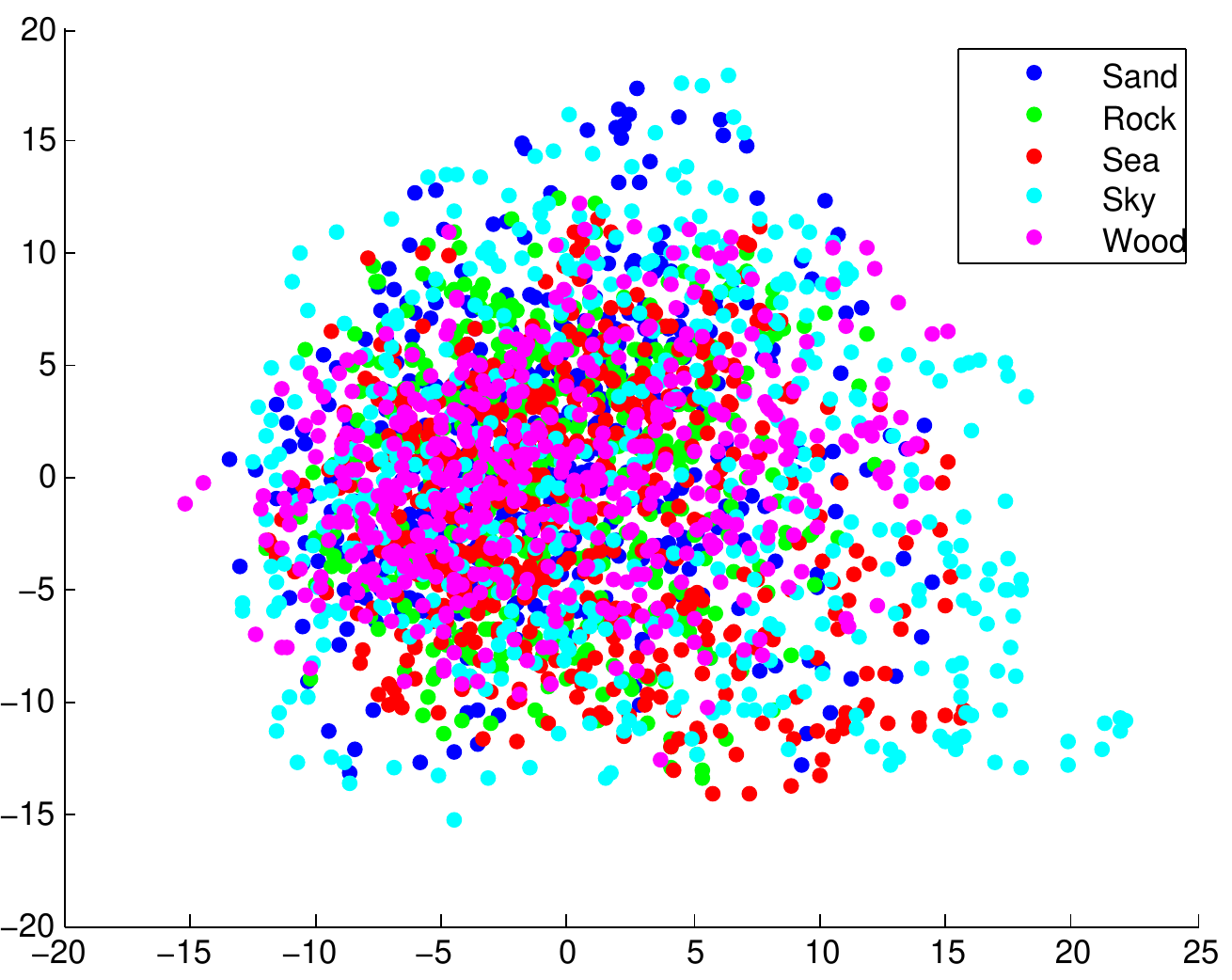}}
    \subfloat[LBP (68.90\%)] {\includegraphics[angle=0, height=0.16\textwidth, width=.2\textwidth]{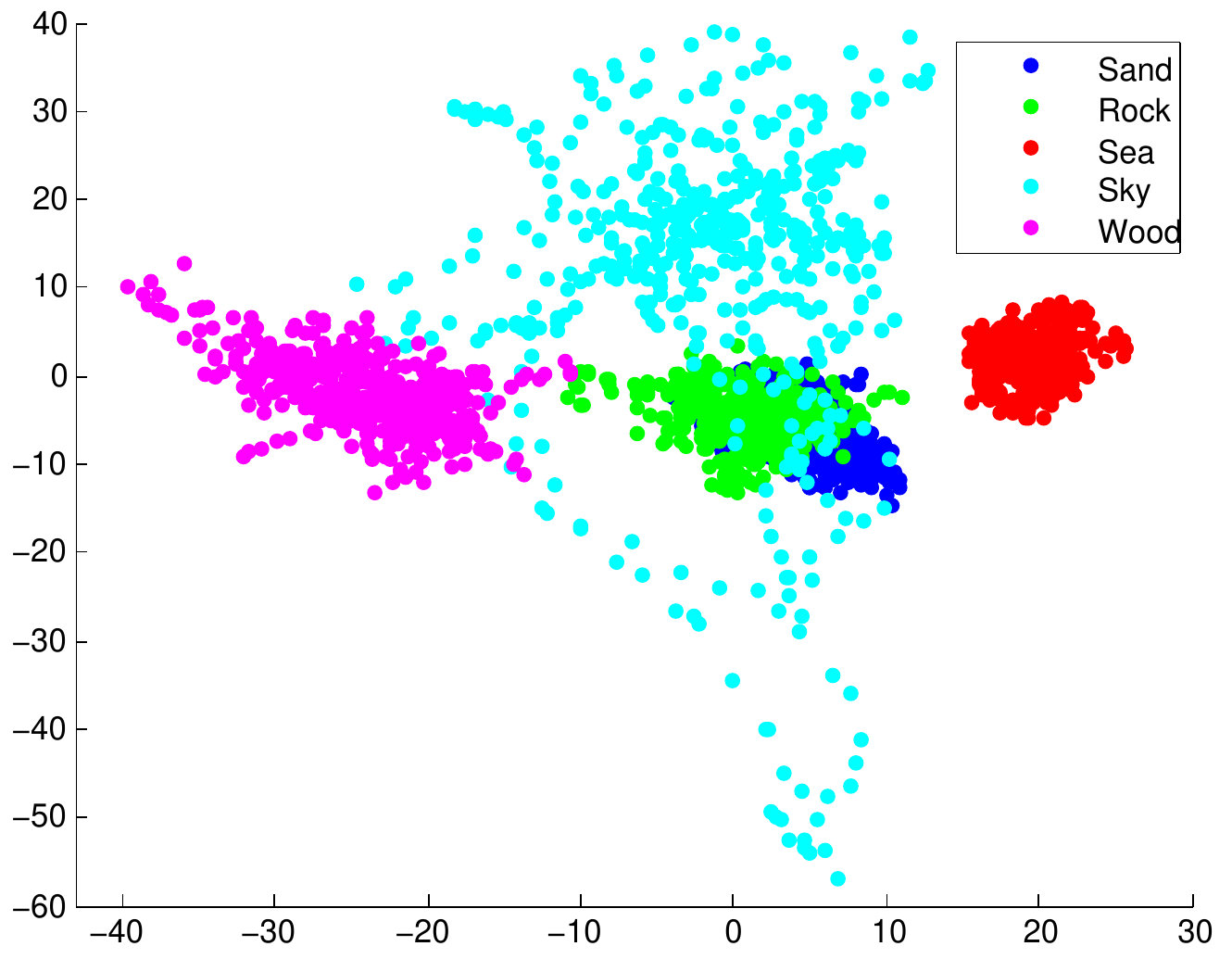}}
\caption{Nature textures represented with different descriptors. For visualization, data are plotted with the dimension reduced to 2 using PCA. Different textures are shown in different colors:
\emph{Sand} (blue), \emph{Rock} (green), \emph{Sea} (red), \emph{Sky} (cyan), \emph{Wood} (magenta).
The k-means clustering accuracies in the parentheses approximately assess the discriminability of each descriptor.
The proposed Weyl descriptor is the most compact and discriminative.
}
\label{fig:natureRep}
\end{figure*}

\begin{figure*} [t]
\centering
 \subfloat[Weyl (\textbf{74.97}\%)] {\includegraphics[angle=0, height=0.16\textwidth, width=.2\textwidth]{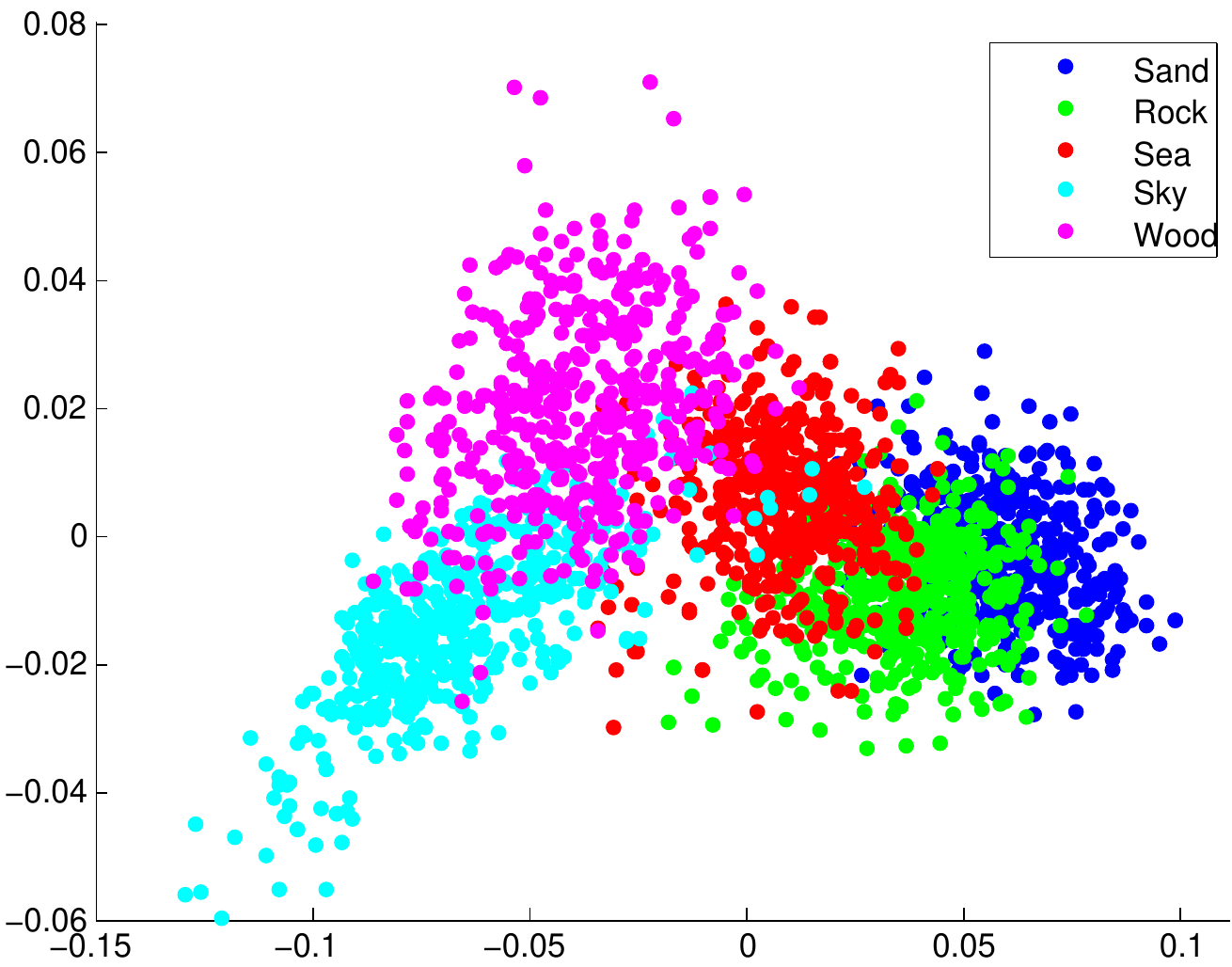} \hspace{0pt}}
  \subfloat[Intensity (31.95\%)] {\includegraphics[angle=0, height=0.16\textwidth, width=.2\textwidth]{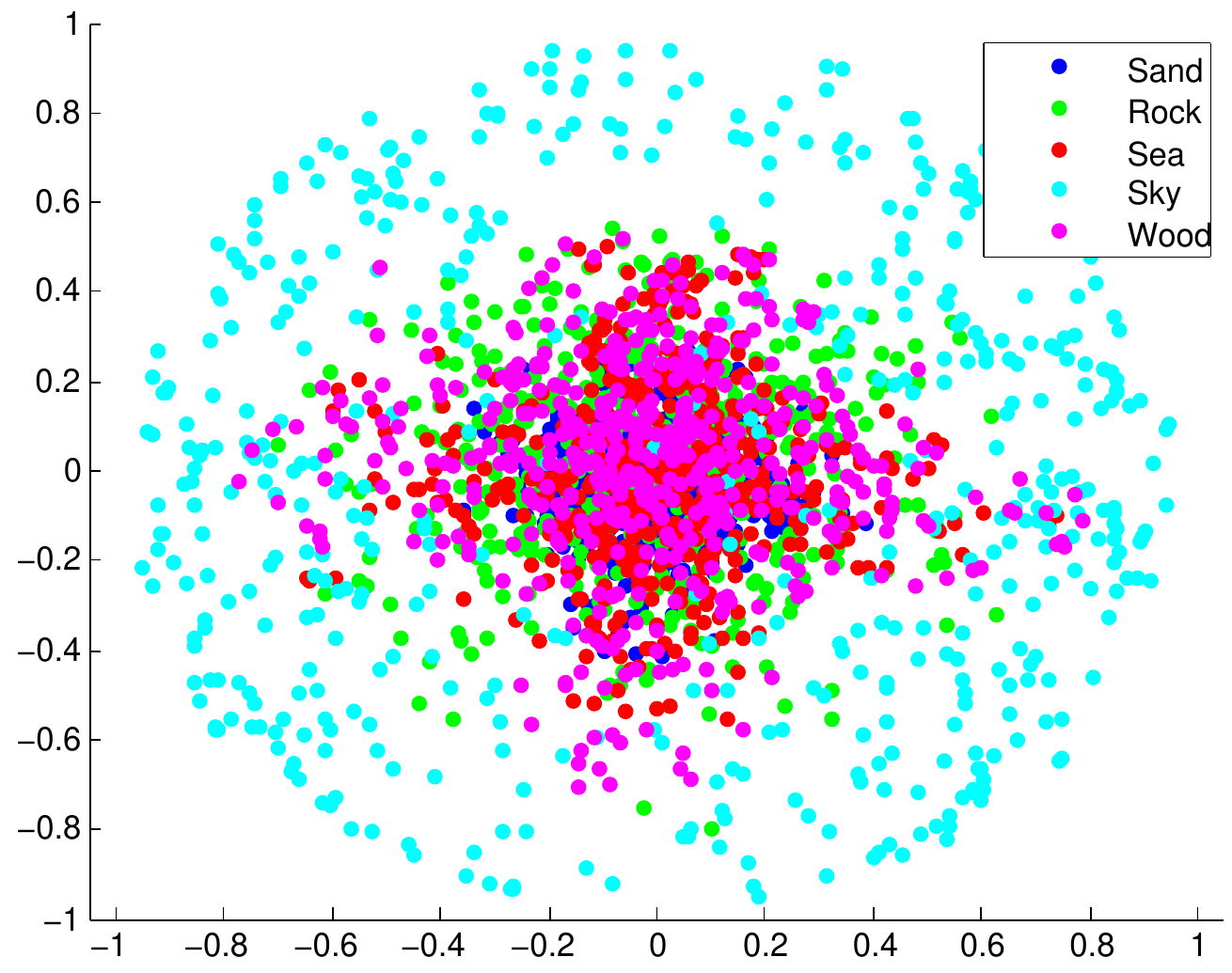}}
    \subfloat[HOG (44.29\%)] {\includegraphics[angle=0, height=0.16\textwidth, width=.2\textwidth]{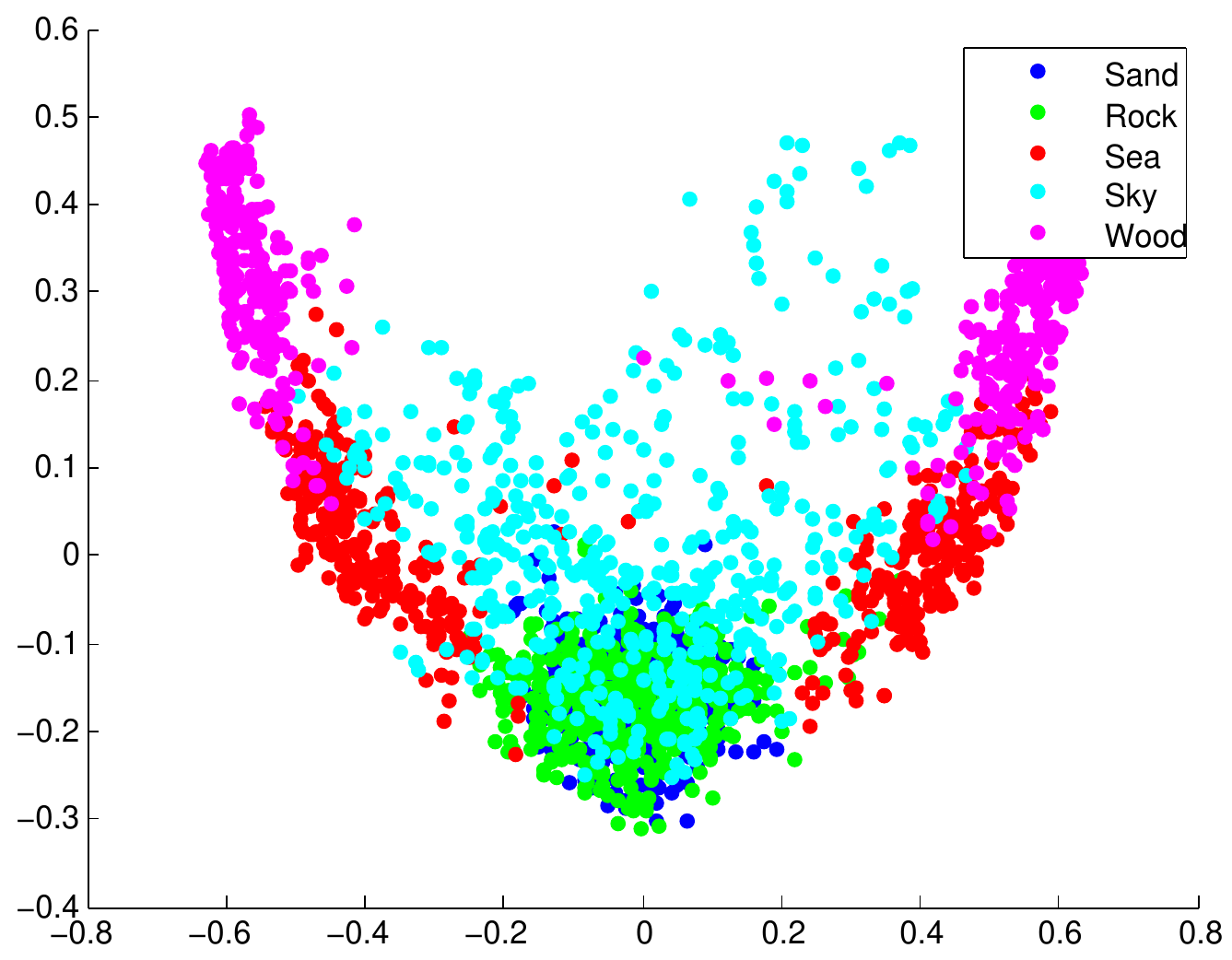}}
  \subfloat[Gabor (26.71\%)] {\includegraphics[angle=0, height=0.16\textwidth, width=.2\textwidth]{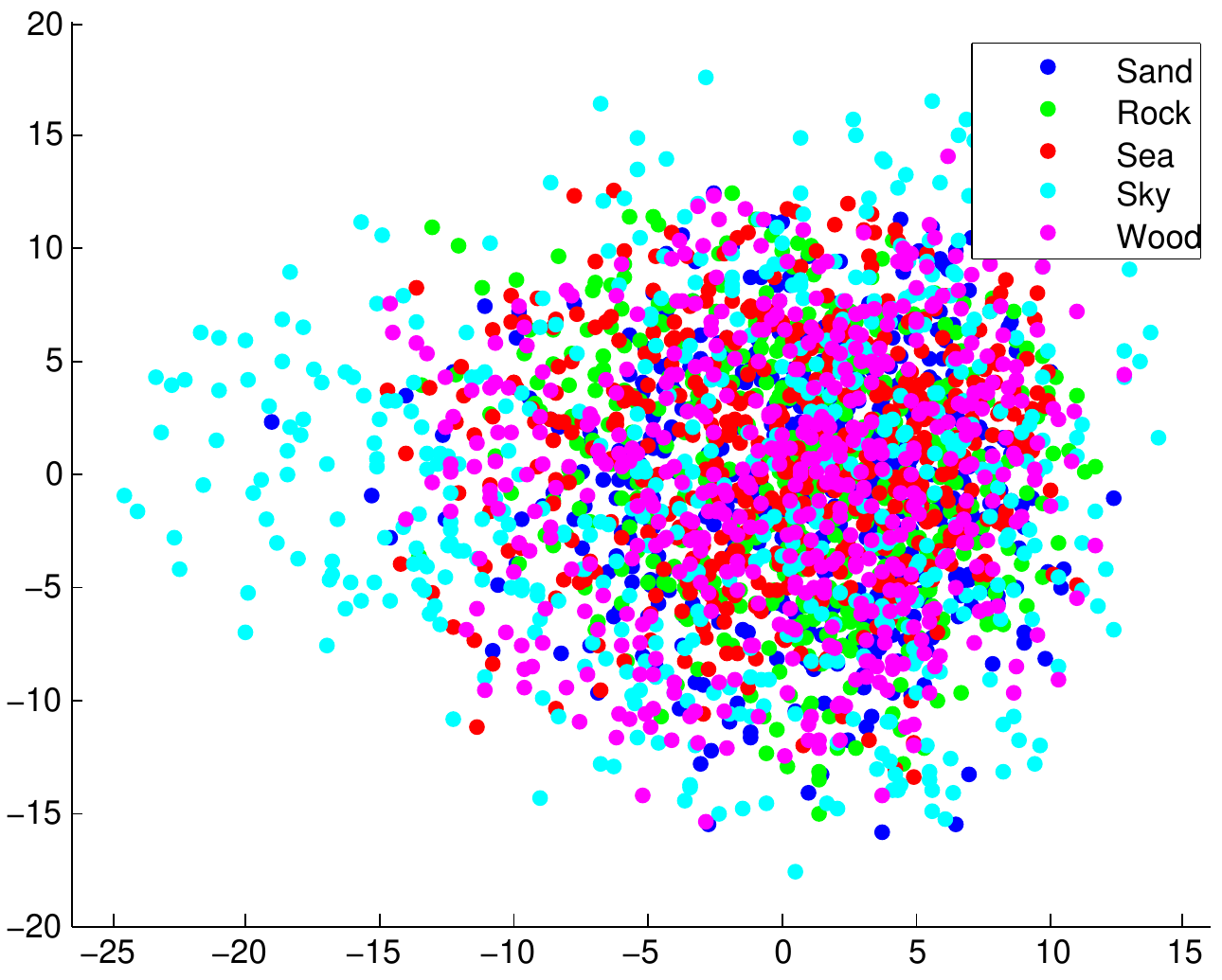}}
    \subfloat[LBP (51.18\%)] {\includegraphics[angle=0, height=0.16\textwidth, width=.2\textwidth]{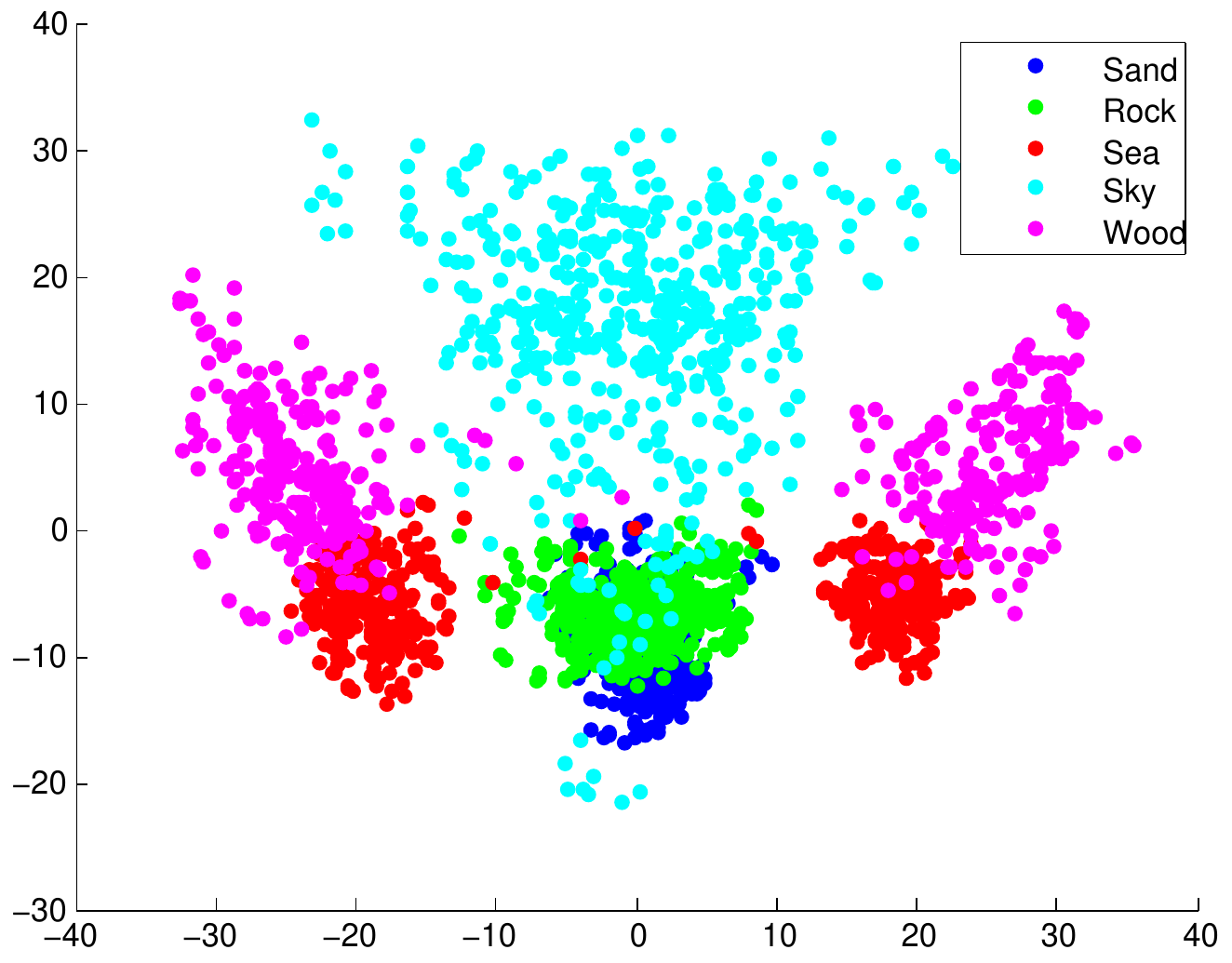}}
\caption{Same as Fig.~\ref{fig:fabricRepDih} for the textures in Fig.~\ref{fig:nature}.
}
\label{fig:natureRepDih}
\end{figure*}

In various texture analysis tasks, e.g., texture classification, segmentation, and retrieval,
it is critical to have a consistent representation for patches of the same texture type; and it is also desirable to have for each patch a small size descriptor to enable efficient computation.
Fig.~\ref{fig:fabricRep} shows the obtained 3500 texture patches represented in different descriptors.
 Different textures are shown in different colors.  For visualization, data are plotted with the dimension reduced to 2 using PCA.
A good texture descriptor encourages a compact and isolated cluster for the same color points. To enable a quantitative assessment, we perform k-means clustering on texture patches represented in different descriptors (before dimension reduction is applied).
The proposed Weyl representation is the most compact and discriminative: a 24-sized descriptor gives 86.49\% clustering accuracy.
 Note that the clustering accuracy here only approximately assesses the discriminability of a representation, since different clustering schemes can lead to different clustering results.
 The popular HOG and LBP provide comparable discriminability, but require an order of magnitude longer descriptors than the Weyl representation.
We notice that the Gabor descriptor fails to effectively represent patches from \emph{Fabric} and \emph{Textile Blue}. After excluding these two textures, we obtain 74.96\% clustering accuracy using Gabor features still below our proposed representation.

We next obtain multiple variants of each texture in Fig.~\ref{fig:fabric} using reflections in horizontal and vertical axes and rotations through multiples of $90^{\circ}$ (dihedral transformations). We repeat the above experiments by sampling 500 patches from multiple variants of each texture. The sampled patches are represented using different descriptors, and plotted in Fig.~\ref{fig:fabricRepDih}
with different colors for different texture types.
By comparing Fig.~\ref{fig:fabricRepDih} to Fig.~\ref{fig:fabricRep}, we observe that the Weyl representation is invariant under dihedral transformations.
For other descriptors,  new sub-clusters emerge among the same color points; in other words,
those descriptors assign significantly different representations to the same texture under different dihedral transformations, causing inferior performance for many texture analysis tasks.

\textbf{Textures from Nature.} We conduct the same set of experiments on textures of significantly less regularity, i.e., the five nature textures in Fig.~\ref{fig:nature}.
As shown in Fig.~\ref{fig:natureRep}, the Weyl representation is still the most compact and discriminative. In Fig.~\ref{fig:natureRepDih}, we again observe the invariance of the Weyl representation under dihedral transformations, enabling significantly more consistent representation than other descriptors.

\subsection{Supervised coefficient selection}

We suggest here another effective way to adopt the Weyl representation through supervised coefficient selection. Consider two-class data points (multi-class cases can be supported using the standard one-vs-all strategy), where the class labels are known beforehand for training purposes. Let ${Y}_+$ and ${Y}_-$ denote the set of points in each of the two classes respectively, where points are arranged as columns of the corresponding matrix. We assume here that the same number of points are selected for each class for computational efficiency.

We now rank all Weyl coordinates based on the discriminability, i.e., how significant they are in distinguishing the classes, and only use the best $K$ Weyl coordinates for classification.
The ability of the Weyl coordinate $(a,b)$ in separating two classes can be assessed by the Weyl coefficient magnitude $|\omega_{a,b}(M)|$ of the matrix ${M}=({Y}_+ - {Y}_-)({Y}_+ + {Y}_-)'$:  We pick one point from each class, $y_+ \in Y_+$ and $y_- \in Y_-$, with their respective Weyl coefficient difference at the coordinate $(a,b)$ given by,
\begin{eqnarray}\label{eq:diff}
&&\left|\mathrm{Tr}[y_+y_+^T D(a,b)]-\mathrm{Tr}[y_-y_-^T D(a,b)]\right|\nonumber\\
&=&\left|\mathrm{Tr}[y_+^T D(a,b)y_+]-\mathrm{Tr}[y_-^T D(a,b)y_-]\right|\nonumber\\
&=&\left|y_+^T D(a,b)y_+ - y_-^T D(a,b)y_-\right|\nonumber\\
&=&\left|(y_+-y_-)^T D(a,b)(y_+ + y_-)\right|\nonumber\\
&=&\left|\mathrm{Tr}[(y_+-y_-)(y_+ + y_-)^T D(a,b)]\right|.
\end{eqnarray}
Thus, we can calculate the total Weyl coefficient difference between $Y_+$ and $Y_-$ at the coordinate $(a,b)$ as $|\mathrm{Tr}[(Y_+-Y_-)(Y_+ + Y_-)^TD(a,b)]|$.
We rank all Weyl coordinates $(a,b)$ in descending order of $|\omega_{a,b}(M)|$, and only use the best $K$ coordinates for classification. Recalling the geometrical interpretation of Weyl coefficients in terms of relative distance from two half-spaces in Section~\ref{sec:HWG}, the aim here is to identity the $K$ subspace pairs whose relative distances discriminate the most between the two classes.

\begin{figure}[!h]
\centering
\includegraphics[angle=0, height=.3\textwidth, width=.39\textwidth]{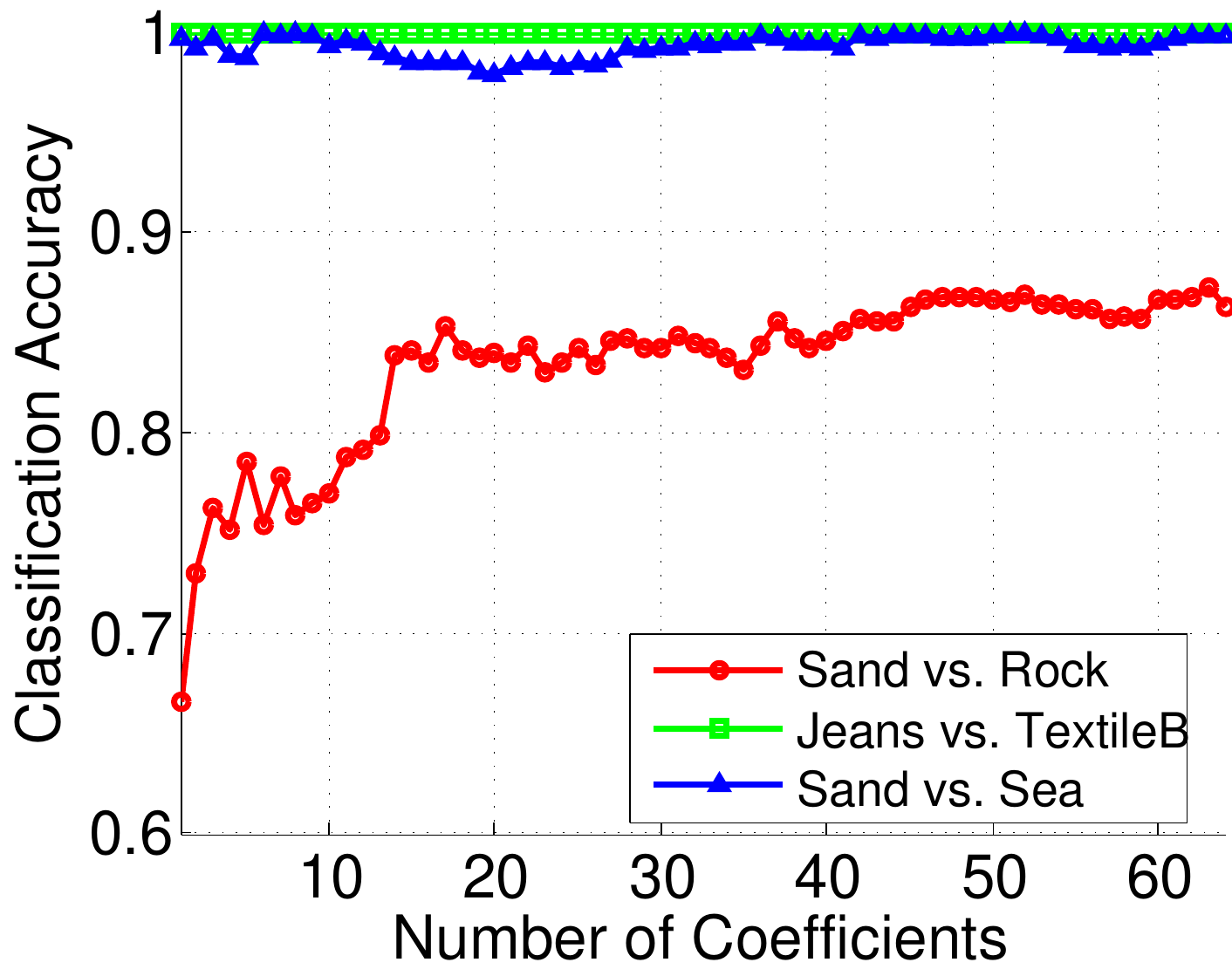}
\caption{Classification accuracies with different number of best Weyl coefficients.}
\label{fig:coeffacc}
\end{figure}

We pick a pair of texture images from Fig.~\ref{fig:fabric} and \ref{fig:nature}, and randomly sample 500 $16\times16$-sized gray-scale patches from each texture. We randomly choose 20 patches for each texture  as training, and the remaining 960 patches as testing.
Patches are represented using different descriptors, and Nearest Neighbor classifiers are used to assign each testing patch to a texture type. Note that, for testing samples, only coefficients at the $K$ selected Weyl coordinates need to be computed.
Table~\ref{tab:coeffacc}  reports the results.
 The best Weyl coefficient already provides accuracies comparable to a 256 long intensity feature descriptor.
When 16 Weyl coefficients are used, we obtain accuracies comparable to all other popular descriptors, but with an order of magnitude shorter descriptor.
Fig.~\ref{fig:coeffacc} shows the classification accuracies when different numbers of best Weyl coefficients are used; note the fast convergence.

\begin{figure}[h]
\centering
{\small
	\begin{tabular}{|l|l||l|l|l|}
	\hline
 & size & Jeans  & Sand  & Sand \\
  &  & vs. TextileB & vs. Sea & vs. Rock\\
	\hline
 \hline
Intensity & 256& 86.45 & 74.06& 66.56\\
HOG & 576& 97.60 & 99.79& 83.54\\
Gabor & 640 & 91.14 & 71.25& 64.79\\
LBP & 256& 100 & 100& 85.41\\
\hline
\hline
Weyl & 1 & 99.89 & 99.68& 66.56 \\
Weyl & 3& 100 & 99.68& 76.25\\
Weyl & 16 & 100 & 99.79& 84.06 \\
\hline
	\end{tabular}
}
\caption{Accuracy (\%) of classifying texture patches represented using different descriptors.
}
\label{tab:coeffacc}
\end{figure}

\section{Proofs}\label{proofs}

This section contains proofs of all results given in Sections~\ref{sec:thm} and~\ref{sec:exa}.


\subsection{Proof of the Weyl transform invariance property}\label{invariance}

Recalling the definitions of the $D(a,b)$ matrices given in (\ref{eq:Dab_def}) and (\ref{eq:XZ_def}), we also introduce the notation
$$e_0:=\left(\begin{array}{c}1\\0\end{array}\right),\;\;\;\;e_1:=\left(\begin{array}{c}0\\1\end{array}\right),$$
for the canonical basis elements in $\RR^2$. In order to prove Lemma 2.1 and Theorem 2.2 we need the following lemma, which gives various properties of the $D(a,b)$ matrices.\\
\\
\begin{lemma}\label{thm:Dab_facts}
Let $X$, $Z$, and $D(a,b)$ be defined in (\ref{eq:Dab_def}) and (\ref{eq:XZ_def}). Then, for all $v,a,b,a',b'\in\ZZ_2^m$,
\begin{flalign}\label{eq:operate}
\mbox{(i)}\;D(a,b)e_v&=(-1)^{b^T v}e_{v+a};&
\end{flalign}
\begin{flalign}\label{eq:transpose}
\mbox{(ii)}\;[D(a,b)]^T&=[D(a,b)]^{-1}=(-1)^{a^T b}D(a,b);&
\end{flalign}
\begin{flalign}\label{eq:mult}
\mbox{(iii)}\;D(a,b)D(a',b')&=(-1)^{a'^T b}D(a+a',b+b').&
\end{flalign}
\end{lemma}

\textbf{Proof:} It follows from (\ref{eq:XZ_def}) that, given $v,a,b\in\ZZ_2$
\begin{equation}\label{eq:XZ_result}
\left(X^a Z^b\right)_v=(-1)^{bv}e_{v+a}.
\end{equation}
Given $v=(v_{m-1}\ldots v_0)^T\in\ZZ_2^m$, we have
\begin{equation}\label{eq:e_kron}
e_{(v_{m-1}\ldots v_0)}=e_{v_{m-1}}\otimes\ldots\otimes e_{v_0},
\end{equation}
which combines with (\ref{eq:Dab_def}) and a standard property of the Kronecker product to give
\begin{eqnarray}
&&D(a,b)e_v\nonumber\\
&=&(X^{a_{m-1}}Z^{b_{m-1}})\otimes\ldots\otimes(X^{a_0}Z^{b_0})(e_{v_{m-1}}\otimes\ldots\otimes e_{v_0})\nonumber\\
&=&(X^{a_{m-1}}Z^{b_{m-1}}e_{v_{m-1}})\otimes\ldots\otimes(X^{a_0}Z^{b_0}e_{v_0}).\label{eq:Dab_kron}\end{eqnarray}
Combining (\ref{eq:XZ_result}) and (\ref{eq:Dab_kron}) then gives
$$\begin{array}{l}
D(a,b)e_v=\\
\left[(-1)^{b_{m-1}v_{m-1}}e_{(v_{m-1}+a_{m-1})}\right]\otimes\ldots\otimes\left[(-1)^{b_0 v_0}e_{(v_0+a_0)}\right]\\
=(-1)^{b^T v}(e_{(v_{m-1}+a_{m-1})}\otimes\ldots\otimes e_{(v_0+a_0)}),\end{array}$$
and (\ref{eq:operate}) now follows by (\ref{eq:e_kron}). It follows from (\ref{eq:operate}) that $[D(a,b)]^2=I$ for all $a,b\in\ZZ_2^m$, from which it follows that $[D(a,b)]^T=[D(a,b)]^{-1}$. Inverting (\ref{eq:operate}), and letting $w=v+a$, we obtain
$$[D(a,b)]^{-1}e_w=(-1)^{b^T(w+a)}e_{w+a}=(-1)^{a^T b}D(a,b)e_w$$
for all $w,a,b\in\ZZ_2^m$, which completes the proof of (\ref{eq:transpose}). We also have, by (\ref{eq:operate}), that
\begin{equation}\label{eq:a_plus_a}
D(a+a',b+b')e_v=(-1)^{(b+b')^T v}e_{v+a+a'}
\end{equation}
for all $v,a,b,a',b'\in\ZZ_2^m$. But two applications of (\ref{eq:operate}) give
\begin{equation}\label{eq:D_product}
\begin{array}{rcl}
D(a,b)D(a',b')e_v&=&D(a,b)(-1)^{b'^T v}e_{v+a'}\\
&=&(-1)^{b^T(v+a')}(-1)^{b'^T v}e_{v+a+a'}\\
&=&(-1)^{a'^T b}(-1)^{(b+b')^T v}e_{v+a+a'}.
\end{array}
\end{equation}
Since (\ref{eq:a_plus_a}) and (\ref{eq:D_product}) hold for all $v\in\ZZ_2^m$, (\ref{eq:mult}) follows.\hfill$\Box$\\
\\
\textbf{Proof of Lemma 2.1:} A symmetric matrix is determined by the entries on and above the diagonal, which gives
\begin{equation}\label{eq:V_dim}
\dim(V)=2^m+\frac{1}{2}\cdot 2^m(2^m-1)=\frac{1}{2}\cdot 2^m(2^m+1).
\end{equation}
On the other hand, writing $a=(a_{m-1}\ldots a_0)^T,b=(b_{m-1}\ldots b_0)^T$, conditioning on the pair $(a_{m-1},b_{m-1})$ leads to the recurrence $|\YY_2^m|=2|\YY_2^m|+2^{2(m-1)}$, from which it is easy to deduce that $|\YY_2^m|=\frac{1}{2}\cdot2^m(2^m+1)$, which combines with (\ref{eq:V_dim}) to give $|\YY_2^m|=\dim(V)$. 

It remains to show that the elements of $\mathcal{B}_{2^m}$ are orthonormal. It follows from (\ref{eq:transpose}) that, for any $(a,b)\in\YY_2^m$,
$$\mathrm{Tr}\left\{\frac{1}{2^{m/2}}[D(a,b)]^T\frac{1}{2^{m/2}}[D(a,b)]\right\}=\frac{1}{2^m}\mathrm{Tr}[(-1)^{a^T b}I]=1,$$
where in the last step we use the assumption that $a^T b=0$. Similarly, it follows from (\ref{eq:transpose}) and (\ref{eq:mult}) that, for any $a,b,a',b'\in\ZZ_2^m$,
\begin{eqnarray}\label{eq:trace_calc}
&&\mathrm{Tr}\left\{\frac{1}{2^{m/2}}[D(a,b)]^T\frac{1}{2^{m/2}}[D(a',b')]\right\}\nonumber\\
&=&\frac{1}{2^m}\mathrm{Tr}\left[(-1)^{a^T b}(-1)^{a'^T b}D(a+a',b+b')\right].
\end{eqnarray}
If $a\neq a'$, $D(a+a',b+b')$ is zero on its diagonal, whereas if $a=0$ and $b\neq 0$, the diagonal terms are nonzero but sum to zero. This observation combines with (\ref{eq:trace_calc}) to show that the elements of $\mathcal{B}_{2^m}$ are orthogonal, which completes the proof.\hfill$\Box$\\
\\
\textbf{Proof of Theorem 2.2} By (\ref{eq:transpose}) and (\ref{eq:mult}), we have, for all $(a,b)\in\YY_2^m$,
\begin{eqnarray*}\label{eq:conj}
&&[D(a',b')]^T D(a,b)D(a',b')\\
&=&(-1)^{a'^T b}[D(a',b')]^T D(a+a',b+b')\\
&=&(-1)^{a'^T b'}(-1)^{a'^T b}D(a',b')D(a+a',b+b')\\
&=&(-1)^{a'^T(b+b')}(-1)^{(a+a')^T b'}D(a,b)\\
&=&(-1)^{a^T b'+a'^T b}D(a,b),
\end{eqnarray*}
which proves the result.\hfill$\Box$

\subsection{Proof of Theorem 2.3}
\label{app:autocorrelation}

We may use (\ref{eq:operate}) to deduce
$$D(a,b)y=\sum_v D(a,b)y_v e_v=\sum_v (-1)^{b^T v}y_v e_{v+a},$$
which gives
\begin{equation}\label{eq:apply_D}
\left\{D(a,b)y\right\}_v=(-1)^{b^T(v+a)}y_{v+a}.
\end{equation}
Meanwhile, the cyclic property of the trace gives
\begin{equation}\label{eq:quad_form}
\mathrm{Tr}[yy^T D(a,b)]=\mathrm{Tr}[y^T D(a,b)y]=y^T D(a,b)y.
\end{equation}
Combining (\ref{eq:apply_D}) and (\ref{eq:quad_form}), we have
\begin{eqnarray}\label{eq:quad_form2}
\{\omega_a(y)\}_b\;\,=\;\,\omega_{a,b}&=&\frac{1}{2^{m/2}}y^T D(a,b)y\nonumber\\
&=&\frac{1}{2^{m/2}}\sum_v y_v\left\{D(a,b)y\right\}_v\nonumber\\
&=&\frac{(-1)^{a^T b}}{2^{m/2}}\sum_v (-1)^{b^T v}y_v y_{v+a}\nonumber\\
&=&\frac{1}{2^{m/2}}\sum_w (-1)^{b^T w}y_w y_{w+a},
\end{eqnarray}
where in the last step we made the substitution $w=v+a$. Now (\ref{eq:quad_form2}) implies that $\{\omega_a(y)\}_b=0$ whenever $a^T b=1$, and the result now follows from the definitions of $H_{2^m}$ and $z_a$.\hfill$\Box$

\subsection{Proof of Proposition 3.1}
\label{app:symplectic}

Rotation by $90^{\circ}$ clockwise can be achieved by an interchange of vertical and horizontal coordinates followed by a reversal of the horizontal coordinates, which is the mapping
\begin{equation}\label{eq:rotation_perm}
\theta:e_{(v_1,v_2)}\longrightarrow e_{(v_2+1_r,v_1)},
\end{equation}
where $y$ is indexed by the binary $2r$-tuple $v=(v_1\;v_2)^T$, $v_1,v_2\in\ZZ_2^r$, and $e_v$ is a canonical basis vector. Writing $a=(a_1\;a_2)^T,b=(b_1\;b_2)^T$, we have by (\ref{eq:rotation_perm}) and (\ref{eq:operate}) that
\begin{eqnarray}\label{eq:rotation_conj}
&&\theta^{-1}D\left(\left[\begin{matrix}a_1\\a_2\end{matrix}\right],\left[\begin{matrix}b_1\\b_2\end{matrix}\right]\right)\theta e_{(v_1,v_2)}\nonumber\\
&=&\theta^{-1}D\left(\left[\begin{matrix}a_1\\a_2\end{matrix}\right],\left[\begin{matrix}b_1\\b_2\end{matrix}\right]\right)e_{(v_2+1_r,v_1)}\nonumber\\
&=&(-1)^{1_r^T b_1+b_1^T v_2+b_2^T v_1}\theta^{-1}e_{(v_2+a_1+1_r,v_1+a_2)}\nonumber\\
&=&(-1)^{1_r^T b_1+b_1^T v_2+b_2^T v_1}e_{(v_1+a_2,v_2+a_1)}.
\end{eqnarray}
Meanwhile, (\ref{eq:operate}) also implies that
$$\begin{array}{rcl}
&&(-1)^{1_r^T b_1}D\left(\left[\begin{matrix}a_2\\a_1\end{matrix}\right],\left[\begin{matrix}b_2\\b_1\end{matrix}\right]\right)e_{v_1,v_2}\\
&=&(-1)^{1_r^T b_1+b_1^T v_2+b_2^T v_1}e_{(v_1+a_2,v_2+a_1)},
\end{array}$$
which combines with (\ref{eq:rotation_conj}) to give the result for rotation. Turning to the translation, consider the permutation
\begin{equation}\label{eq:translation_perm}
t:e_{(v_1,p,q,v_2)}\longrightarrow e_{(v_1,p+q,q+1,v_2)},
\end{equation}
where $v_1\in\ZZ_2^r$, $v_2\in\ZZ_2^{r-2}$and $p,q\in\ZZ_2$. The mapping $t$ permutes the first two vertical binary coordinates as
$$\left[\begin{array}{cccc}
0&0&1&1\\0&1&0&1\end{array}\right]\longrightarrow\left[\begin{array}{cccc}0&1&1&0\\1&0&1&0\end{array}\right],$$
which gives a vertical cyclic translation through $N/4$. Writing $a=(a_1\;j\;k\;a_2)^T,b=(b_1\;l\;m\;b_2)^T$, we have by (\ref{eq:translation_perm}) and (\ref{eq:operate}) that
\begin{equation}\label{eq:translation_conj}
\begin{array}{l}
t^{-1}D\left(\left[\begin{smallmatrix}a_1\\j\\k\\a_2\end{smallmatrix}\right],\left[\begin{smallmatrix}b_1\\l\\m\\b_2\end{smallmatrix}\right]\right)t e_{(v_1,p,q,v_2)}=\\
t^{-1}D\left(\left[\begin{smallmatrix}a_1\\j\\k\\a_2\end{smallmatrix}\right],\left[\begin{smallmatrix}b_1\\l\\m\\b_2\end{smallmatrix}\right]\right)e_{(v_1,p+q,q+1,v_2)}=\\
(-1)^{b_1^T v_1+b_2^T v_2+l(p+q)+m(q+1)}t^{-1}e_{(v_1+a_1,p+q+j,q+1+k,v_2+a_2)}\\
=(-1)^{b_1^T v_1+b_2^T v_2+lp+lq+mq+m}e_{(v_1+a_1,p+j+k+1,q+k,v_2+a_2)}.\end{array}
\end{equation}
Meanwhile, (\ref{eq:operate}) also implies that
$$\begin{array}{l}(-1)^{m}D\left(\left[\begin{smallmatrix}a_1\\j+k\\k\\a_2\end{smallmatrix}\right],\left[\begin{smallmatrix}b_1\\l\\l+m\\b_2\end{smallmatrix}\right]\right)e_{v_1,p,q,v_2}\\
=(-1)^{b_1^T v_1+b_2^T v_2+lp+lq+mq+m}e_{(v_1+a_1,p+j+k+1,q+k,v_2+a_2)},\end{array}$$
which combines with (\ref{eq:translation_conj}) to give the result for translation.\hfill$\Box$

\section{Concluding remarks}
\label{sec:con}

We have described how to represent measurement using the binary Weyl transform and developed new theory, supported by texture classification experiments, that connects mathematical properties of the transform with a broad class of signal processing objectives. In particular, we used invariance of the transform to a large class of multi-scale dihedral transformations to pool Weyl coefficients, thereby developing very concise histograms with high discriminative power.  We also described a supervised method that learns the transform coefficient with the greatest discriminative value through training. Illustrative examples using real world texture examples were provided for both approaches.

{
\bibliographystyle{IEEEtran}
\bibliography{weyl}
}

\end{document}